\documentclass[twoside]{article}
\usepackage[accepted]{aistats2025}
\usepackage{amsmath,amsfonts,bm}

\def\eqref#1{equation~\ref{#1}}

\def\1{\bm{1}}

\DeclareMathAlphabet{\mathsfit}{\encodingdefault}{\sfdefault}{m}{sl}
\SetMathAlphabet{\mathsfit}{bold}{\encodingdefault}{\sfdefault}{bx}{n}

\newcommand{\E}{\mathbb{E}}

\newcommand{\R}{\mathbb{R}}

\newcommand{\Var}{\mathrm{Var}}

\newcommand{\bA}{{\bf A}}
\newcommand{\bO}{{\bf O}}
\newcommand{\bL}{{\bf \Lambda}}

\newcommand{\bS}{{\bf S}}

\newcommand{\PP}{\mathbb P}
\renewcommand{\tilde}{\widetilde}

\newtheorem{theorem}{Theorem}[section]

\newtheorem{proposition}[theorem]{Proposition}
\newtheorem{lemma}[theorem]{Lemma}

\usepackage[utf8]{inputenc} 
\usepackage[T1]{fontenc}    
\usepackage{hyperref}       
\usepackage{booktabs}       
\usepackage{amsfonts}       
\usepackage{nicefrac}       
\usepackage{microtype}      
\usepackage{xcolor}         
\usepackage{multirow}
\usepackage{graphicx}
\usepackage{subcaption}
\usepackage{cleveref}
\usepackage{natbib}
\usepackage{authblk}
\usepackage{mathrsfs}
\usepackage{bbm}
\usepackage{comment}
\usepackage{breqn}
\usepackage{hyperref}
\usepackage{url}
\usepackage{algorithm}
\usepackage{algorithmic}
\usepackage{relsize}
\usepackage{xparse}
\usepackage{xspace}

\newcommand{\pa}{\partial}

\begin{document}

%

%
\runningauthor{X. Liu, H. Du, W. Deng, R. Zhang}

\twocolumn[

\aistatstitle{Optimal Stochastic Trace Estimation in Generative Modeling}

\aistatsauthor{ Xinyang Liu$^{*}$ \And Hengrong Du$^{*}$ \And  Wei Deng \And Ruqi Zhang}

\aistatsaddress{ Purdue University \And  UC Irvine \And Morgan Stanley \And  Purdue University}]

\begin{abstract}
Hutchinson estimators are widely employed in training divergence-based likelihoods for diffusion models to ensure optimal transport (OT) properties. However, this estimator often suffers from high variance and scalability concerns. To address these challenges, we investigate Hutch++, an optimal stochastic trace estimator for generative models, designed to minimize training variance while maintaining transport optimality. Hutch++ is particularly effective for handling ill-conditioned matrices with large condition numbers, which commonly arise when high-dimensional data exhibits a low-dimensional structure. To mitigate the need for frequent and costly QR decompositions, we propose practical schemes that balance frequency and accuracy, backed by theoretical guarantees. Our analysis demonstrates that Hutch++ leads to generations of higher quality. Furthermore, this method exhibits effective variance reduction in various applications, including simulations, conditional time series forecasts, and image generation. The code is available at \url{https://github.com/xinyangATK/GenHutch-plus-plus}.
\end{abstract}


\section{Introduction}

Diffusion models and score-based generative models (SGMs) \citep{DDPM, score_sde} have been widely used in areas such as text-to-image synthesis, video generation, and audio synthesis \citep{text_2_image, imagen_video, DiffWave}. The simulation-free diffusion process results in a simple and elegant loss function, which enhances training stability, flexibility, and interoperability, and significantly improves scalability. However, diffusion models are known to suffer from weak transport properties \citep{Lavenant_Santambrogio_22} and require many function evaluations to generate the desired data.

SGMs draw inspiration from score matching \citep{score_matching}, which was originally developed to address energy-based models. However, the requirement to compute the trace of the model's density renders this approach unscalable in relation to data dimensions. To mitigate this challenge, \cite{Vincent_2011} transformed score matching into a denoising problem, thus circumventing the need for trace computation. Additionally, \cite{estimate_Hessian_curvature} proposed a rank-1 approximation of the Hessian by backpropagating curvature. For more efficient trace estimators, methods such as sliced score matching and FFJORD \citep{FFJORD, song2020sliced} employed the Hutchinson estimator \citep{Hutchinson89} to enhance computation by leveraging deterministic approximation of large eigenvalues and Hessian-vector products and enables more efficient transportation plans in diffusion studies.

Since the introduction of the Hutchinson estimator, significant progress has been made in diffusion models with optimal transport (OT) guarantees. For instance, the Trajectory Net (trajNet) \citep{TrajectoryNet} established a connection between continuous normalizing flows (CNF) \citep{neural_ode} and dynamic optimal transport \citep{Benamou_Brenier_2020}. Furthermore, the Schrödinger bridge \citep{forward_backward_SDE, mSB, SBP_max_llk, reflected_schrodinger_bridge, VSDM, VSMD} explored a principled framework for dynamic optimal transport \citep{DSB} by training divergence-based objectives. However, despite the advantages offered by optimal transport properties, the high variance associated with the Hutchinson estimator has posed scalability challenges, making it less favorable in real-world generative modeling tasks.

To address the large-variance issue directly while maintaining transport optimality, we study variance reduction techniques to minimize the variance of the Hutchinson estimator without resorting to simulation-free solutions \citep{flow_matching, Albergo_stochastic_interpolants}. Hutchinson estimators are known to yield reasonable approximations with a flat spectrum. However, this approach contradicts empirical evidence showing that generated high-dimensional data lies in a manifold of low intrinsic dimension \citep{fefferman2016testing, diffusion_manifold, Manifolds_Hypothesis_image}. Therefore, we propose leveraging Hutch++, the optimal stochastic trace estimator \citep{hutch_pp}, to compute large eigenvalues in a deterministic manner to reduce variance and approximate the rest using the Hutchinson estimator. We demonstrated the effectiveness of the proposed method across a wide range of generative models, including neural ODEs \citep{neural_ode, FFJORD}, sliced score matching \citep{song2020sliced}, TrajNet \citep{TrajectoryNet}, and Schrödinger-based diffusion models \citep{forward_backward_SDE, mSB, SBP_max_llk, VSDM, VSMD}. Our method also has the potential to be applied to other generative models.
We summarize our contributions in four aspects:

\begin{itemize}
    \item We propose Hutch++ estimators to optimally minimize variance, thereby improving the trace estimation of a large class of diffusion models with optimal transport guarantees.
    \item We identify that conducting matrix decompositions too frequently incurs high costs in generative modeling. We propose a practical algorithm that amortizes the decompositions to reduce costs.
    \item We provide theoretical guarantees to demonstrate the higher fidelity of the generated content using Hutch++ estimators.  
    \item The effectiveness of our proposed models is tested in simulations, time series forecasts, and image generation in both conditional and unconditional settings.  
\end{itemize}

\section{Related Works}
\paragraph{Simulation-free Samplers} \cite{flow_matching} accelerated the training of continuous normalizing flows (CNFs) in a simulation-free manner, motivating the development of straighter probability paths with more efficient transport. \citet{Albergo_stochastic_interpolants, Albergo_unified_framework} bridged the gap between diffusion models and flow models by stochastically interpolating the marginals. \cite{Rectified, Rectified_group} proposed the rectified flow, which iteratively rectifies couplings and probability paths. \citet{Multisample_flow_matching,CFM_Tong} studied optimal transport in big data settings using minibatch optimal transport. However, these methods primarily rely on convex transport costs, and extending them to nonlinear cases and developing a more principled approach to updating couplings remain open questions.

\paragraph{Optimal Transport} \citet{DSB, SBP_max_llk, forward_backward_SDE, mSB} explored Schrödinger bridge diffusion models, a principled framework for studying optimal transport (OT). However, these models have been known to scale poorly to large-scale datasets. Inspired by flow matching and diffusion models, \cite{SB_matching, gSBM} investigated bridge matching to develop more scalable methodologies while preserving efficient transport maps. \cite{NLSB, NOT_LC} studied Lagrangian formulations to ensure more directional trajectories, especially when obstacles exist in the transport dynamics. Nonetheless, achieving optimality often incurs additional costs, especially the large variance issue, and finding ways to  reduce the training variance remains a longstanding challenge.

\section{Preliminaries}
\subsection{Implicit Loss in Neural ODE}
The flow-based generative models uses invertible function $f:\mathbb{R}^D \rightarrow \mathbb{R}^D$ to map points from prior to data distribution~\citep{density_real_NVP, dinh2014nice}. 
Given a latent variable $\mathbf{z}\sim q(\mathbf{z})$, the parameterized model $p_\theta(x)$ can be evaluated as:
\begin{equation}\label{eq: change_of_variables}
    \log p_\theta(\mathbf{x}) = \log q(\mathbf{z}) - \log \det \left| \frac{\partial f(\mathbf{z})}{\partial \mathbf{z}} \right|
\end{equation}
where $\partial f(\mathbf{z}) / \partial \mathbf{z}$ is the Jacobian of $f$.
To avoid $\mathcal{O}(D^3)$ cost in determinant computation, 
\citet{neural_ode} presented continuous normalizing flows, mapping the reverse process through ordinary differential equations (ODEs) defined by the parameterized function $f(\mathbf{z}(t),t;\theta)$. Following 
~\citet{neural_ode}, 
Equation \ref{eq: change_of_variables} can be re-formulated in integral form:
\begin{equation}\label{eq: ode_likelihood}
    \log p_\theta(\mathbf{z}(t_1)) = \log q(\mathbf{z}(t_0)) - \int_{t_0}^{t_1} \mathrm{Tr} \left( \frac{\partial f}{\partial \mathbf{z}(t)} \right) \, dt.
\end{equation}
where $\mathbf{z}(t_0) = \mathbf{z_0}$ is sampled from a  initial distribution $q(\mathbf{z_0})$ and $\mathbf{z}(t_1) = \mathbf{x}$ denotes the observed samples. 

\subsection{Hutchinson Trace Estimator}\label{sec: Hutchinson}
Given a jacobian matrix $\mathbf{A} \in \mathbb{R}^{D \times D}$, the exact trace $\mathrm{Tr}(\mathbf{A})$ requires computing the full Jacobian matrix,
which is prohibitive in practice when $D$ increases.
Fortunately, the Hutchinson trace estimator~\citep{Hutchinson89} is a computational technique used to approximate the trace of a large matrix efficiently. 
The method first involves generating random vector $\mathbf{v}_i \in \mathbb{R}^D$, and then computes the product $\mathbf{v}_i^\top \mathbf{A} \mathbf{v}_i$ for each vector. 
Thus, we can get an unbiased estimate of the trace:
\begin{equation}\label{eq: hutchinson_unbaised}
    \mathrm{Tr}(\mathbf{A}) = \E_{p(\mathbf{v})}[\mathbf{v}^\top\mathbf{A}\mathbf{v}]
\end{equation}
for all random variable satisfied $\E_{p(\mathbf{v})}[\mathbf{v}^\top\mathbf{v}] = I$. 
Therefore, the trace can be estimated by Monte Carlo using Hutchinson's trace estimator~\citep{Hutchinson89}:
\begin{equation}\label{eq: hutchinson}
    \mathrm{Tr}(\mathbf{A}) \approx \frac{1}{m} \sum_{i=1}^{m} \mathbf{v}_i^\top \mathbf{A} \mathbf{v}_i=:H_m(\mathbf{A}),
\end{equation}
where each $\mathbf{v}_i$ are $i.i.d.$ samples from $p(\mathbf{v})$, typically from a Gaussian distribution~\citep{Hutchinson89}. 
$m$ denotes the number of matrix-vector multiplication queries~\citep{avron2011matrixvector1, roosta2015matrixvector2}.
Leveraging the Hutchinson estimator, \citet{FFJORD} proposed FFJORD, a normalizing flow model trained with a free-form Jacobian, which extends the potential for training divergence-based likelihood in generative modeling~\citep{FFJORD,song2020sliced, forward_backward_SDE}.

\subsection{Likelihood Training of Schrödinger Bridge Models}\label{sec: fb-sde}
The dynamic Schrödinger bridge (SB) problem is defined as solving:
\begin{equation}
    \min_{\mathbf{Q} \in \mathcal{P}(p_{\text{data}}, p_{\text{prior}})} D_{KL}(\mathbf{Q} \parallel \mathbf{P}),
\end{equation}
where ${\mathbf{Q} \in \mathcal{P}(p_{\text{data}}, p_{\text{prior}})}$ denotes a set of \textit{path measures} with marginals probability density $p_{\text{data}}$ and $p_{\text{prior}}$ at $t=0$ and $t=T$, respectively. $\mathcal{P}$ is a prior measure typically defined via Brownian motion.
To solve the SB problem, \citet{forward_backward_SDE} proposed divergence-based likelihood training of Forward-Backward Stochastic Differential Equations (FB-SDE), a framework transforms the optimality condition of SB into a set of SDEs:
\begin{align}
\scalebox{0.95}{$
    \mathrm{d}\mathbf{x}_t = \left[f(\mathbf{x}_t, t) + \beta_t \nabla_{\mathbf{x}} \log \overrightarrow{\psi}_t(\mathbf{x}_t, t)\right]\mathrm{d}t + \sqrt{\beta_t}\mathrm{d}\mathbf{w}_t $}\tag{6a}\label{eq: fb-sde-b}\\
    \scalebox{0.95}{$
    \mathrm{d}\mathbf{x}_t = \left[f(\mathbf{x}_t, t) - \beta_t \nabla_{\mathbf{x}} \log \overleftarrow{\varphi}_t(\mathbf{x}_t, t)\right]
    \mathrm{d}t + \sqrt{\beta_t}\mathrm{d}\mathbf{w}_t $}\tag{6b}\label{eq: fb-sde-f}
\end{align}
where $f(\cdot, t):\R^d\rightarrow\R^d$ and $\beta_t$ denote the vector field and time-varying scalar; $\mathbf{x}_0 \sim p_{\text{data}}$ and $\mathbf{x}_0 \sim p_{\text{prior}}$. 
$\nabla_{\mathbf{x}} \log \overleftarrow{\psi}_t(\mathbf{x}_t, t)$ and $\nabla_{\mathbf{x}} \log \overrightarrow{\varphi}_t(\mathbf{x}_t, t)$ are the optimal forward and backward drifts for SB.
$\mathbf{w}_t$ is standard Wiener process.
Given the models $\overleftarrow{z}_t^\theta$ and $\overrightarrow{z}_t^\phi$, we can learn the forward and backward policies $\overleftarrow{z}_t = \sqrt{\beta_t}\nabla_{\mathbf{x}} \log \overleftarrow{\psi}_t$ and $\overrightarrow{z}_t = \sqrt{\beta_t}\nabla_{\mathbf{x}} \log \overleftarrow{\varphi}_t$, respectively. 
We list two simplified divergence-based objectives of SB model in an alternate training scheme:
\begin{align}
\scalebox{0.85}{$
    \Tilde{\mathcal{L}}_{SB}(\mathbf{x}_0) = - \int_{0}^{T}\E_{\mathbf{x}_t\sim \ref{eq: fb-sde-b}}\left[ \frac{1}{2}\|\overrightarrow{z}_t^\phi\|^2 + \sqrt{\beta_t}\nabla_{\mathbf{x}} \cdot\overrightarrow{z}_t^\phi + \overleftarrow{z}_t^\top\overrightarrow{z}_t^\phi\right] \mathrm{d}t
    $} \tag{7a}\label{eq: fb-sde-train-b}\\
\scalebox{0.85}{$
    \Tilde{\mathcal{L}}_{SB}(\mathbf{x}_T) = - \int_{0}^{T}\E_{\mathbf{x}_t\sim \ref{eq: fb-sde-f}}\left[ \frac{1}{2}\|\overleftarrow{z}_t^\theta\|^2 + \sqrt{\beta_t}\nabla_{\mathbf{x}} \cdot\overleftarrow{z}_t^\theta + \overrightarrow{z}_t^\top\overleftarrow{z}_t^\theta\right] \mathrm{d}t 
$}\tag{7b}\label{eq: fb-sde-train-f}
\end{align}
where the divergence terms in Equation \ref{eq: fb-sde-train-b} and Equation \ref{eq: fb-sde-train-f} can be approximated by Hutchinson estimator \citep{Hutchinson89}.
The divergence-based likelihood training paradigm enables the Schrödinger bridge model to use the forward policy $\overleftarrow{z}_t$ to build the optimal path governing samples towards $p_{\text{prior}}$ with optimal transport guarantee.
This addressed the weak transport properties in SGM and enriches its architecture by introducing the nonlinear drifts into the learning process under the guarantee of optimal transport.

\section{Optimal Stochastic Trace Estimation in Generative Modeling}
Introducing the Hutchinson estimator in the dynamics solver~\citep{FFJORD} provides an unbiased log-density estimation with $\mathcal{O}(D)$ cost, allowing more flexible model architectures, as well as developing the divergence-based likelihood training paradigm in diffusion models~\cite{song2020sliced, forward_backward_SDE, VSDM, provably_schrodinger_bridge}.
However, the Hutchinson estimator leads to high variance, which gradually accumulates over the duration of each integration, posing a scalability challenge in model training. 
In this section, we will present a variance reduction technique to mitigate this issue and integrate it with modern diffusion models, enhancing its applicability for generative modeling tasks.

\subsection{Variance Reduced Divergence-Based Training} \label{sec: variance_reduction}
As discussed in~\citet{hutch_pp}, the Hutchinson estimator is less sensitive to individual large eigenvalues, which implies that matrices with a flatter spectrum can be approximated more reasonably. 
However, in real-world generative modeling scenarios, high-dimension data often exhibits underlying patterns captured by fewer dimensions~\citep{van2008tsne, ng2011sparse, VAE, carlsson2009topology}.
Motivated by the desirable properties of the low-rank approximation,
we proposed leveraging Hutch++~\citep{hutch_pp} as a natural variance reduced version of Hutchinson estimator. Hutch++ involves a low-rank approximation step and estimate $ \mathrm{Tr} \left( \frac{\partial f_t}{\partial \mathbf{z}(t)} \right)$ through two parts:
\begin{itemize}
    \item deterministic computation of large eigenvalues;
    \item stochastic estimation of small eigenvalues.
\end{itemize}
We elaborate on the Hutch++ estimator as follows.

Given $i.i.d.$ random vectors $S_i \in \R^{D}$ and $G_i \in \R^{D}$, typically from a Gaussian distribution. Hutch++ stars by computing an orthonormal span $Q$ through a single iteration of the power method with a random initial vector $S$.
Here $Q_i \in \R^D$ provides an approximate estimation of $i$-th top eigenvector of $\mathbf{A}$. 
Then we can separate $\mathbf{A}$ into two parts: the projection onto the subspace spanned by $Q$ and the projection onto the orthogonal complement of this subspace. 
Therefore, we can divide the calculation of the $\mathrm{Tr}(\mathbf{A})$ into two parts:
\setcounter{equation}{7}
\begin{equation}\label{eq: trance_separate}
    \mathrm{Tr}(\mathbf{A}) = \mathrm{Tr}(Q^\top\mathbf{A}Q)+ \mathrm{Tr}((I-QQ^\top)\mathbf{A}(I-QQ^\top))
\end{equation}
where $Q = \mathrm{QR}(\mathbf{A}S)$ and $\mathrm{QR}(\cdot)$ performs QR decomposition and return the orthonormal basis.
The first term is derived from $\mathrm{Tr(QQ^\top\mathbf{A}QQ^\top)}$ by leveraging the cyclic property of the trace and can be computed exactly. 
Consequently, the error in estimating $\mathrm{Tr}(\mathbf{A})$ arises solely from the approximation of the second term. This term is approximated using the Hutchinson estimator with the random vector $G_i$, resulting in a significantly lower variance in the estimation of $\mathrm{Tr}((I-QQ^\top)\mathbf{A}(I-QQ^\top))$ compared to the direct estimation of $\mathrm{Tr}(\mathbf{A})$ in Equation \ref{eq: ode_likelihood}.
We present the method for trace estimation using Hutch++ \citep{hutch_pp}:
\begin{equation}
\scalebox{0.87}{$
    H_m^{++}(\mathbf{A})=\mathrm{Tr}(Q^\top\mathbf{A}Q)+H_{\frac{m}{3}}((I-QQ^\top)\mathbf{A}(I-QQ^\top)).$}\label{eqn:hutchpp} 
\end{equation}
To enhance the scalability of the diffusion model for high-dimensional data while preserving transport efficiency, we adapt the proposed method to Schrödinger bridge (SB)-based diffusion models~\citep{forward_backward_SDE, mSB, VSDM}, which are alternately trained with divergence-based likelihood introduced in \Cref{sec: fb-sde}.
For the divergence-based training objectives, 
we propose leveraging the Hutch++ estimator to approximate the divergence terms in Equation \ref{eq: fb-sde-train-b} \ref{eq: fb-sde-train-f}. 

Similar to~\citep{forward_backward_SDE, VSDM, FFJORD}, we sample each initial vector $S_i$ and noise vector $G_i$ independently and fix them over the integration interval. We re-formulate Equation \ref{eq: fb-sde-train-b} as:
\begin{dmath}\label{eqn:loglike2}
    \Tilde{\mathcal{L}}_{SB}(\mathbf{x}_0) = - \int_{0}^{T}\E_{\mathbf{x}_t\sim \ref{eq: fb-sde-b}}\left[ \frac{1}{2}\|\overrightarrow{z}_t^\phi\|^2 + \overleftarrow{z}_t^\top\overrightarrow{z}_t^\phi\\
    \scalebox{0.98}{$+ \sqrt{\beta_t}\mathrm{Tr}\tiny{\left(\frac{\partial \overrightarrow{z}_t^\phi}{\partial \mathbf{x}(t)}\right)}$}\right] \mathrm{d}t \\
    = - \int_{0}^{T}\E_{\mathbf{x}_t\sim \ref{eq: fb-sde-b}}\left[ \frac{1}{2}\|\overrightarrow{z}_t^\phi\|^2 + \overleftarrow{z}_t^\top\overrightarrow{z}_t^\phi\\
    + \scalebox{1.0}{$\sqrt{\beta_t}\left(\mathrm{Tr}\left(Q^\top\frac{\partial \overrightarrow{z}_t^\phi}{\partial \mathbf{x}(t)}Q\right) 
    + \mathrm{Tr}\left(P^\perp\frac{\partial \overrightarrow{z}_t^\phi}{\partial \mathbf{x}(t)}P^\perp\right)\right)$}\right] \mathrm{d}t \\
    = - \int_{0}^{T}\E_{\mathbf{x}_t\sim \ref{eq: fb-sde-b}}\left[ \frac{1}{2}\|\overrightarrow{z}_t^\phi\|^2 + \overleftarrow{z}_t^\top\overrightarrow{z}_t^\phi \\
    + \scalebox{0.86}{$\sqrt{\beta_t}\left(\mathrm{Tr}\left(Q^\top\frac{\partial \overrightarrow{z}_t^\phi}{\partial \mathbf{x}(t)}Q\right)
    + \E_{p(G)} \left[ G^\top\left(P^\perp \frac{\partial \overrightarrow{z}_t^\phi}{\partial \mathbf{x}(t)}P^\perp \right)G\right]\right)$}\right] \mathrm{d}t
\end{dmath}
where $Q = \mathrm{QR}(\frac{\partial f_t}{\partial \mathbf{z}(t)}S)$ and $P^\perp = I - QQ^\top$ denotes the orthogonal complement projection matrix of $Q$.
Likewise, the forward training stage can be improved with reduced variance facilitated by the Hutch++ estimator.

From the perspective of probability flow, $\Tilde{\mathcal{L}}_{SB}(\mathbf{x}_0)$ used in the backward training stage collapses to the log-likelihood in Equation \ref{eq: ode_likelihood} when the drift degenerates. 
Consequently, SB models can be articulated through the concept of flow within the flow-based model training framework~\citep{forward_backward_SDE, song2019generative, gong2021interpreting}, thereby enhancing the scalability of flow-based models while improving transport efficiency.

\vspace{-1mm}
\subsection{Acceleration Technique for Training}\label{sec: acceleration}

\vspace{-1mm}
For SB-based diffusion models, leveraging Hutch++ to estimate $\mathrm{Tr}\left(\frac{\partial z_t}{\partial \mathbf{x}(t)}\right)$\footnote{Hereafter, We use $\frac{\partial z_t}{\partial \mathbf{x}(t)}$ to represent the Jacobian matrix in log-likelihood of SB-based diffusion models} reduced the high variance introduced by the Hutchinson estimator, leading to faster convergence in training and higher quality density estimation.
However, as the data dimension and batch size increase, updating $Q$ via QR decomposition on $\frac{\partial z_t}{\partial \mathbf{x}(t)} S$ at each step over a single integral interval becomes prohibitively time-consuming.
To tackle this challenge, we propose acceleration techniques for training divergence-based likelihood under the Hutch++ estimator, where we update the orthogonal matrix $Q$ every $L_s$ steps over the duration of each integration.

Inspired by the principle of efficient communication in federated learning~\citep{mcmahan2017fedlearning1}, 
we focus on a few updates of $Q$, ensuring each update takes on more responsibility, thereby reducing the overall number of updates.
With the observation of the noising/denoising process in diffusion models,
we regard the change in $\frac{\partial f_t}{\partial \mathbf{z}(t)}$ across a continuous period of time to be minor, indicating that the top eigenvalues of $\frac{\partial f_t}{\partial \mathbf{z}(t)}S$ during this interval are approximately the same. This motivates us to share the $Q$ required for subsequent trace estimation within this continuous time interval.
Specifically, we divide the integral interval with a total of $L$ steps into multiple subintervals, each consisting of $L_s$ steps.
The orthogonal matrix $Q^t_i$ of each step can be mathematically defined as: for $t_0=0$ and $t_L=1$\footnote{For clarity, we denotes the diffusion interval as $[t_0, t_L]$.}
\begin{align}
\scalebox{0.95}{$
    Q^{t_0} =  \text{QR}\left(\frac{\partial z_t}{\partial \mathbf{x}(0)} S\right), \quad  Q^{t_L} =  \text{QR}\left(\frac{\partial z_t}{\partial \mathbf{x}(1)}S\right) $}\notag \hspace{0.95cm}\\
\scalebox{0.95}{$
    Q^{t_i} = \text{QR}\left(\frac{\partial z_t}{\partial \mathbf{x}\left(\lfloor\frac{L}{L_s} t_i\rfloor \cdot \frac{L_s}{L}\right)}S\right),\quad \text{for}\quad t\in[t_i, t_{i+L_s}) $}\label{eqn:acceleration}
\end{align}
where $\lfloor \cdot\rfloor $ denotes the floor function.
With the help of shared orthogonal matrices $Q^t$, we only need to perform QR decomposition about $\lfloor\frac{L}{L_s}\rfloor$ times over the duration of each integration, instead of updating $Q^t$ every step.
This technique significantly accelerates the training process under divergence-based likelihood for diffusion models.
We provide a theoretical analysis in \Cref{sec: error_propagation}, along with an experimental comparison on synthetic data in~\Cref{sec: exp_simulation}.

\section{Theoretical Analysis}

\subsection{Preliminaries}

The Hutchinson trace estimator provides an efficient means to estimate the trace of a matrix, with the following properties:

\begin{lemma}[\cite{Hutchinson89}]
\begin{equation}
    \mathbb{E}[H_m(\mathbf{A})] = \mathrm{Tr}(\mathbf{A}), \; \mathrm{Var}[H_m(\mathbf{A})] \le \frac{2}{m} \mathrm{Tr}^2(\mathbf{A}), \label{eqn:VanHut}
\end{equation}
where $H_m(\mathbf{A})$ is the Hutchinson estimator with $m$ random vectors, $\mathbb{E}[\cdot]$ denotes expectation, and $\mathrm{Var}[\cdot]$ denotes variance.
\end{lemma}

To reduce the variance further, especially for matrices with large dominant eigenvalues, we can leverage a low-rank QR approximation. The Hutch++ estimator improves upon the original Hutchinson method by significantly reducing the variance:

\begin{lemma}[\cite{hutch_pp}] 
Suppose $\mathbf{A}$ is a \emph{symmetric positive semidefinite (PSD)} matrix. Then, the Hutch++ estimator satisfies:
\begin{equation}
    \mathbb{E}[H_m^{++}(\mathbf{A})] = \mathrm{Tr}(\mathbf{A}),\mathrm{Var}[H_m^{++}(\mathbf{A})] \le \frac{18}{m(m-3)} \mathrm{Tr}^2(\mathbf{A}),
\end{equation}
where $H_m^{++}(\mathbf{A})$ is the Hutch++ estimator.
\end{lemma}

It is worth noting that the PSD assumption can be generalized to estimates involving the nuclear norm of non-PSD matrices. However, in this paper, we focus on matrices where large eigenvalues are the primary concern, and thus the PSD assumption always holds.

We define the \emph{relative error} of an estimator $T(\mathbf{A})$ as:
\begin{equation*}
    \varepsilon(T) := \frac{|T(\mathbf{A}) - \mathrm{Tr}(\mathbf{A})|}{|\mathrm{Tr}(\mathbf{A})|}.
\end{equation*}

A key advantage of Hutch++ is its enhanced variance reduction, which leads to significant computational savings compared to the original Hutchinson estimator. Specifically:

\begin{proposition}[\cite{hutch_pp}]
    For a given error threshold $\varepsilon > 0$ and confidence level $\delta > 0$, the following holds with probability at least $1 - \delta$:
    \begin{itemize}
        \item The error of the Hutchinson estimator satisfies $\varepsilon(H_m) < \varepsilon$ when $m = \mathcal{O}\Big(\frac{\log(1/\delta)}{\varepsilon^2}\Big)$.
        \item The error of the Hutch++ estimator satisfies $\varepsilon(H_m^{++}) < \varepsilon$ when $m = \mathcal{O}\Big(\sqrt{\frac{\log(1/\delta)}{\varepsilon^2}} + \log(1/\delta)\Big)$.
    \end{itemize}
\end{proposition}

This result highlights that Hutch++ achieves a quadratic reduction in the required number of samples $m$ for a given accuracy, compared to the original Hutchinson estimator.

\subsection{Error Analysis for Acceleration}
Naïvely applying Hutch++ in generative modeling can incur significant computational costs due to the need for QR decompositions at every iteration. To mitigate this, we propose updating the QR decomposition every $L_s$ iterations. While this approach means the estimated eigenvalues may not always be fully up to date, it strikes a balance between computational efficiency and accuracy.
We now analyze the errors introduced by perturbations in the matrix
$\mathbf{A}$. Let $\widetilde{\mathbf{A}}$ be a perturbed version of
$\mathbf{A}$, by the smoothness of the dynamics, the difference of their traces is proportional to a small
time increment $\eta$ and the frequency of steps in QR updates $L_s$, i.e.,$\mathrm{Tr}(\widetilde{\mathbf{A}}) =\mathrm{Tr}(\mathbf{A}) +
\mathcal{O}((L_s-1)\eta)$. Consider the matrix $Q = \mathrm{QR}(\mathbf{A}S)$,
where $S$ is a random sketching matrix. The \emph{approximate Hutch++ estimator}
for this perturbed matrix is given by:
\begin{equation*}
\scalebox{0.9}{$
    \widetilde{H}_m^{++}(\widetilde{\mathbf{A}}) := \mathrm{Tr}(Q^\top \widetilde{\mathbf{A}} Q) + H_{\frac{m}{3}}\left( (I - QQ^\top) \widetilde{\mathbf{A}} (I - QQ^\top) \right).$}
\end{equation*}

This approximation results from reducing the frequency of QR updates, which is central to our acceleration method. For example, using the frozen QR decomposition (Equation \ref{eqn:acceleration}) over the time subinterval $[t_i, t_{i+L_s}]$, $\frac{\partial z_t}{\partial \mathbf{x}}$ behaves as a perturbation of $\frac{\partial z_t}{\partial \mathbf{x}\left(\lfloor\frac{L}{L_s} t_i\rfloor \cdot \frac{L_s}{L}\right)}$.

We justify the effectiveness of this estimator by proving the following expectation and variance bounds:
\begin{proposition}\label{prop:approximateHut}
\begin{align}
\scalebox{0.9}{$
    \mathbb{E}[\widetilde{H}_m^{++}(\widetilde{\mathbf{A}})] = \mathrm{Tr}(\widetilde{\mathbf{A}})$},\hspace{3.35cm}\tag{14}\label{eqn:approximateExp}\\
\scalebox{0.9}{$
    \mathrm{Var}(\widetilde{H}_m^{++}(\widetilde{\mathbf{A}})) \le \frac{36}{m(m-3)} \mathrm{Tr}^2(\mathbf{A}) +  \frac{1}{m}\mathcal{O}((L_s-1)^2\eta^2).$}\tag{15}\label{eqn:approximateVar}
\end{align}
\end{proposition}
In Equation \ref{eqn:approximateExp}, we observe that the approximate Hutch++ estimator provides the exact trace of $\tilde{\mathbf{A}}$, demonstrating its robustness to the QR decomposition used in the acceleration process. Additionally, in Equation \ref{eqn:approximateVar}, the approximate Hutch++ estimator achieves an approximately quadratic variance reduction similar as the standard Hutch++ estimator.  
Conducting QR decomposition only every \(L_s\) iterations reduces computations while maintaining a variance reduction comparable to the vanilla version. Regarding the specific guidance on the selection of \(L_s\), we refer to interested readers to the complexity analysis section 1.5 in the supplementary file.

\subsection{Error Propagation in Divergence-Based Likelihood Training}\label{sec: error_propagation}

The accumulation of error during the training of divergence-based likelihoods can be analyzed as follows. In the case of Neural ODEs \cite{neural_ode}, by applying the frozen QR decomposition (Equation \ref{eqn:acceleration}) over the time subinterval \( [t_i, t_{i+L_s}] \) for Equation \ref{eq: ode_likelihood}, we estimate the trace term, assuming that the approximate Hutch++ operator is adapted to \( \frac{\partial f}{\partial \mathbf{z}} \) within this interval. Let \( \log \tilde{p}_\theta(t) \) denote the resulting log-likelihood under this approximation. Using Proposition \ref{prop:approximateHut}, we can then establish the following result:

\setcounter{equation}{15}
\begin{proposition}
Let \( M = \max\{|\mathrm{Tr}(\frac{\partial f}{\partial \mathbf{z}})| : t \in [0, T]\} \). Then for \( t \in [0, T] \), we have:
\begin{equation}
\scalebox{0.95}{$
\begin{aligned}
  \mathbb{E}\left[ \log \tilde{p}_\theta(t) \right] &= \log p_\theta(\mathbf{z}(t)),\\
  \mathrm{Var}[\log \tilde{p}_\theta(t)] &\leq T^2 \left[\frac{36M^2}{m(m-3)} + \frac{1}{m}\mathcal{O}((L_s-1)^2\eta^2)\right].
\end{aligned}\label{eqn:loglikelihoodest}$}
\end{equation}
\end{proposition}

Since the divergence-based likelihood of the Schrödinger bridge \citep{forward_backward_SDE, mSB} differs from neural ODEs only by an additional deterministic inner product (see Equation \ref{eq: fb-sde-train-f}) vs. the ISM loss in \cite{VSDM}), the variance reduction achieved by Hutch++ estimators in Equation \ref{eqn:loglikelihoodest} naturally extends to Schrödinger bridge (SB)-based diffusion models \citep{forward_backward_SDE, mSB, VSDM} as well.

\section{Experiments}
We test our method on a variety of generative modeling tasks including density estimation on simulations, time series imputation and forecasting, as well as image data modeling. 
Our baselines include various Neural ODE model variants and divergence-based Schrödinger bridge models. We describe the details of the implementation in Appendix 2.2.

\subsection{Simulations}\label{sec: exp_simulation}
We first apply the proposed Hutch++ estimator to the Neural ODE model FFJORD \citep{FFJORD}. We refer to our enhanced version as FFJORD++.
We validated FFJORD++ on several toy distributions that serve as standard benchmarks \citep{FFJORD, wehenkel2019simulations1}. 
In this section, we note that the model is evaluated using exact trace when estimating density or reporting the test loss (negative log-likelihood) of the test set during training.

\paragraph{Analysis on Training Efficiency}
\begin{figure}[!t]
    \centering
    \begin{subfigure}[b]{0.48\textwidth}
        \centering
        \includegraphics[width=1.\textwidth]{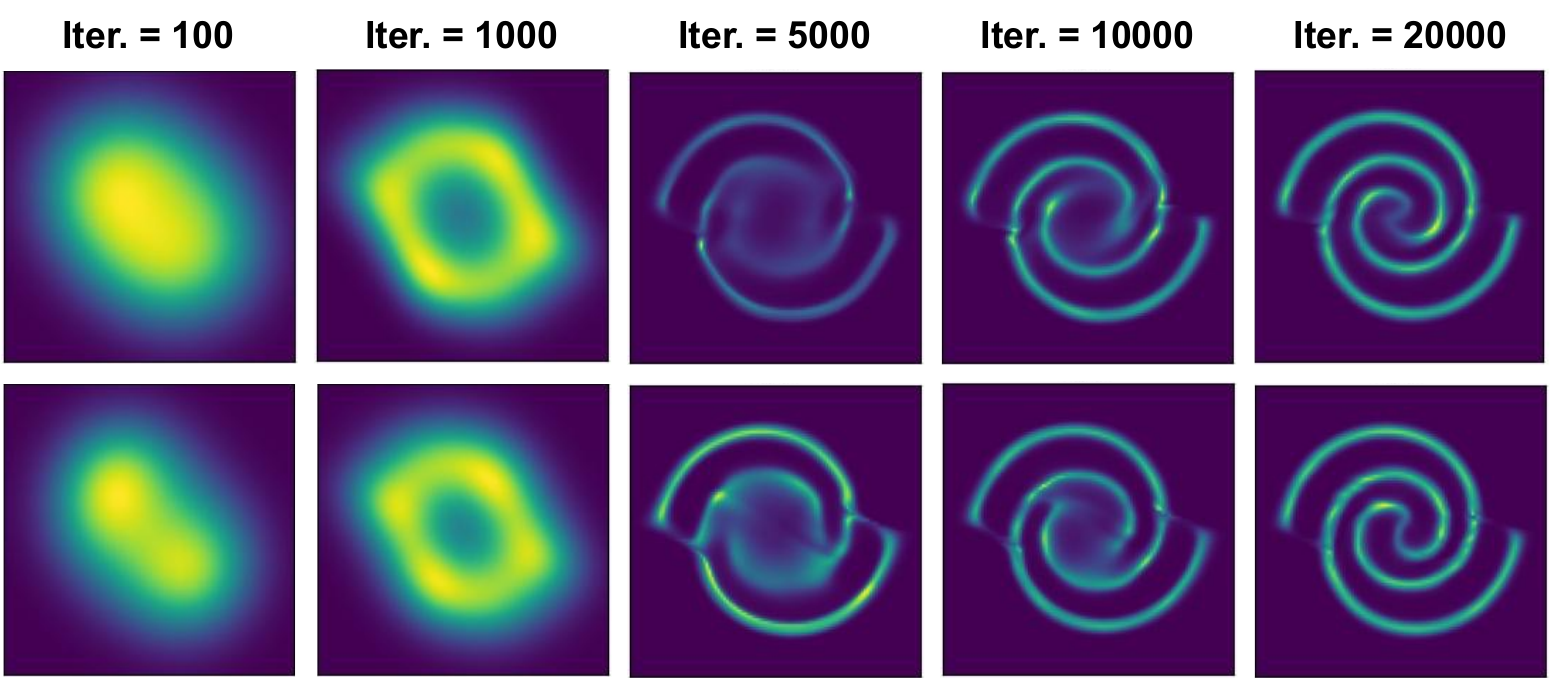}
        \label{fig: 2s}
    \end{subfigure}
    \vspace{-5mm}
    \caption{\small Visualization of density estimation obtained by \textbf{(Top) FFJORD}  and \textbf{(Bottom) FJORD++} during the training phase on \texttt{2spirals}. The proposed variance-reduced model, FJORD++, improves both convergence and training stability, resulting in higher-quality estimated densities.}
    \vspace{-3mm}
    \label{fig: vis_train}
\end{figure}

\begin{figure*}[!t]
    \centering
    \begin{subfigure}[b]{0.9\textwidth}
        \centering
        \includegraphics[width=\textwidth]{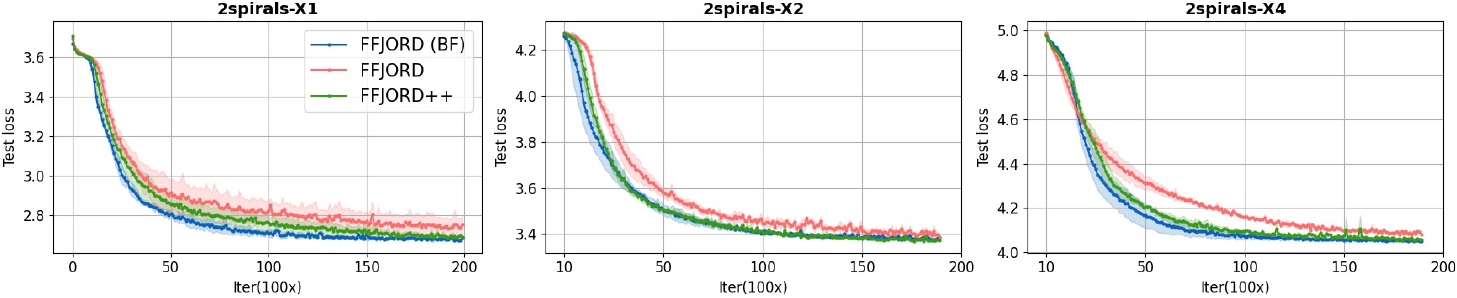}
        \label{fig: scale_x1}
        \vspace{-4mm}
    \end{subfigure}
    \vspace{-3mm}
    \caption{\small Comparison between the FFJORD++ and FFJORD on \texttt{2spirals} with various scales. \textbf{From left to right}, each row represents the data with a default scale, as well as shapes stretched 2x and 4x along the X-axis, respectively. As the scale along the X-axis increases, FFJORD++ consistently exhibits superior convergence rates, further widening the performance gap with FFJORD.}
    \vspace{-3mm}
    \label{fig: condition_num}
\end{figure*}
Figure \ref{fig: vis_train} visualizes the comparison of the density estimated by FFJORD and FFJORD++ during the training stage on \texttt{2spirals}. 
Visually, the density estimated by FFJORD++ exhibits more distinct modes in the early stages and achieves improved stability in subsequent phases compared to the density estimated by FFJORD. 
This observation suggests that FFJORD++ has the potential to fit data distributions faster and recover data distributions more accurately, exhibiting the advantages of employing Hutch++ to reduce variance in Equation \ref{eq: ode_likelihood} in model training. We provide more examples in the Appendix 3.1.

\paragraph{Density Modeling with Different Condition Number}
To better validated the applicability of our method, we further explored its performance in fitting general shapes.
In addition to the default scale, we stretch the X-axis of the data by 2 and 4 times and make comparisons between FFJORD ++ and FFJORD on \texttt{2spiral}
with various scales in \Cref{fig: condition_num}.
We denote them by 2spiral-1X, 2spiral-2X and 2spiral-4X, respectively.
As shown in \Cref{fig: condition_num}, FFJORD++ exhibits faster convergence rates than FFJORD and even achieves lower testing loss, which is comparable to results training with exact trace. 
This observation suggests that the variance reduction technique in \Cref{sec: variance_reduction} can improve the convergence and stability of model training, resulting in higher-quality estimated densities.
Furthermore, 
as the scale along the X-axis is stretched, FFJORD++ consistently demonstrates superior convergence rates, further widening the performance gap with FFJORD.
This observation indicates that FFJORD++ is superior for modeling data distributions with higher condition numbers caused by increased stretching scales, which in turn impacts the eigenvalue distribution of the $\frac{\partial f_t}{\partial \mathbf{z}(t)}$ during training. 
We explain this to FFJORD++ can better approximate the $ \mathrm{Tr} \left( \frac{\partial f_t}{\partial \mathbf{z}(t)} \right)$ in Equation \ref{eq: ode_likelihood} by projecting the top eigenvalues when the matrix's spectrum is not flat, aligning with our estimate (Equation \ref{eqn:loglikelihoodest}). We provide more results in the Appendix 3.1.

\begin{figure}[!t]
    \centering
    \begin{subfigure}{0.5\textwidth}
        \centering
        \hspace{-0.7cm}
        \includegraphics[width=0.8\linewidth]{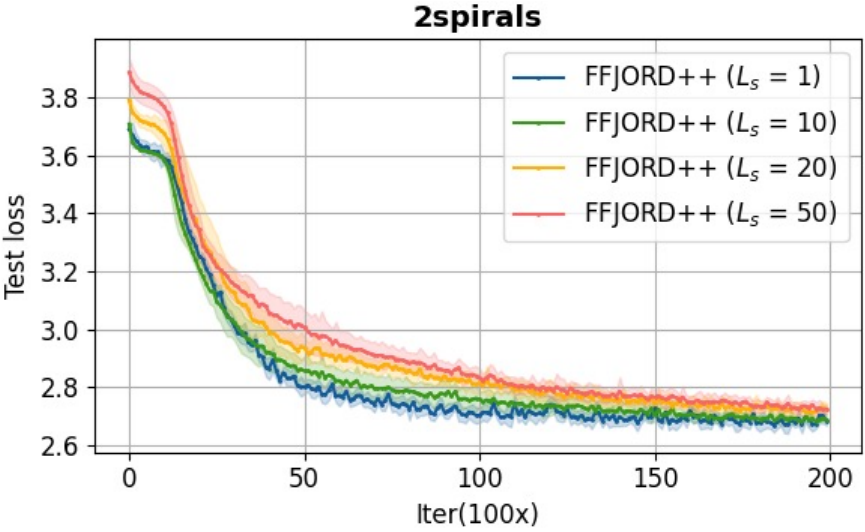}
    \end{subfigure}
    \begin{subfigure}{0.5\textwidth}
        \centering
        \vspace{2mm}
        \hspace{-0.5cm}
        \scalebox{0.8}{
        \begin{tabular}{l|c|cccc}
        \toprule
        Method & FFJORD &\multicolumn{4}{c}{FFJORD++} \\
         &  & S=1 & S=10 & S=25 & S=50\\
        \midrule
        Cost (second) &38 &407 &55 &50 &43\\
        \bottomrule
        \end{tabular}}
        \label{tab: cost}
    \end{subfigure}
    \label{fig: shared_step}
    \caption{\small \textbf{(Top)} Comparison of test loss during training on FFJORD++ trained with different $L_s$. \textbf{(Left)} Time cost for 100 iterations on models trained with different $L_s$.}
    \label{fig: ablation}
    \vspace{-6mm}
\end{figure}

\paragraph{Ablation on Shared Step Size $L_s$}
As described in \Cref{sec: acceleration}, there exists a trade-off between computational cost and variance when estimating the trace within integration in Equation \ref{eq: ode_likelihood}, where the lower variance can lead to faster convergence during training.
Therefore, we ablate the shared steps $L_s$ on 2spirals. Specifically, the default step size for the entire integration interval is 100 and we update the orthogonal matrix $Q$ each $L_s$ steps, where $L_s \in [1, 10, 25, 50]$. We show the comparison of test loss during training on FFJORD++ trained with different $L_s$ in \Cref{fig: ablation} \textbf{(Top)}.
As shown in~\Cref{fig: ablation} \textbf{(Bottom)}, we compare the computational cost of FFJORD and various versions of FFJORD++ with shared steps $L_s$ during training, over 100 iterations with batch size set to 512. 
Considering the results \Cref{fig: ablation}, it is evident that vanilla FFJORD++ ($L_s$=1) enhances model convergence.
However, its computational cost is roughly ten times higher than that of FFJORD, making it impractical for real-world applications.
Fortunately, increasing the length of shared subintervals, that is, reducing the count of computing $Q$ over the duration of each integration, does not significantly degrade the model's performance. 
Instead, it yields in substantial time savings, making the approach much more efficient.
Based on the above observations, we chose $L_s = 10$ as the default shared steps. 
The improvements introduced by FFJORD++ and its competitive time efficiency compared to FFJORD motivate us to further extend Hutch++ to accommodate a broader range of data types and scenarios.

\subsection{Time Series Modeling}
\begin{table*}[!t]
    \centering
    \caption{\small \textbf{(Left)} Results on Time Series Imputation. \textbf{(Right)} CRPS on Time Series Forecasting. (lower is better).}
        \label{tab: exp_time}
    \begin{minipage}{0.5\textwidth} 
        \centering
        \scalebox{0.65}{
        \begin{tabular}{lccccccccc}
        \toprule[1.5pt]
        &\multicolumn{3}{c}{PM2.5} &\multicolumn{3}{c}{PhysioNet 0.1} &\multicolumn{3}{c}{PhysioNet 0.5} \\
        \cmidrule(lr){2-4}\cmidrule(lr){5-7}\cmidrule(lr){8-10}
        Methods
        & RMSE 
        & MAE
        & CRPS
        & RMSE 
        & MAE
        & CRPS
        & RMSE 
        & MAE
        & CRPS\\
        \midrule
        V-RIN & 40.1& 25.4& 0.53&0.63 &0.27 &0.81 &0.69 &0.37 &0.83\\ 
        Multitask GP &42.9 &34.7 &0.41 &0.80 &0.46 &0.49 &0.84 &0.51 &0.56\\
        GP-VAE &43.1 &26.4 &0.41 &0.73 &0.42 &0.58 &0.76 &0.47 &0.66\\
        CSBI &20.7 &10.8 &\textbf{0.11} &0.62 &0.29 &0.35 &0.72 &0.35 &0.39\\
        \textbf{CSBI++(Ours)} &\textbf{20.2} &\textbf{10.3} &\textbf{0.11} &\textbf{0.54} &\textbf{0.24} &\textbf{0.28} &\textbf{0.66} &\textbf{0.33} &\textbf{0.34}\\
        \bottomrule
    \end{tabular}}
    \end{minipage}
    \hspace{1.5cm}
    \begin{minipage}{0.4\textwidth} 
        \centering
        \label{tab: tsi}
        \scalebox{0.78}{
        \begin{tabular}{lccc}
        \toprule[1.5pt]
        Methods & Exchange & Solar &Electricity\\
        \midrule
        GP-copula &0.008 &\textbf{0.317} &0.041 \\
        Vec &0.009 &0.384 &0.043 \\
        TransMAF &0.012 &0.368 &0.039 \\
        CSBI &0.008 &0.372 &0.038 \\
        \textbf{CSBI++ (Ours)} &\textbf{0.007} &0.363 &\textbf{0.035} \\
        \bottomrule
    \end{tabular}}
    \end{minipage}
\end{table*}

To validate the effectiveness of our method on real-world data, we applied Hutch++ to the cutting-edge model CSBI~\citep{provably_schrodinger_bridge}, 
a conditional Schrödinger bridge method for time series modeling.
Our enhanced method is referred to as CSBI++, and we conducted experimental quantitative analysis through time series imputation and forecasting.

\paragraph{Datasets and Metrics}\label{sec: exp_time_data}
For the imputation task, we run experiments on two datasets. 
The air quality dataset \texttt{PM2.5}~\citep{yi2016pm25} and the healthcare dataset \texttt{PhysioNet} in PhysioNet Challenge 2012~\citep{silva2012physio}.
For \texttt{PhysioNet}, we randomly mask a specific percentage (10\%, 50\%) of the data that comprise the test dataset, denoted as \texttt{PhysioNet 0.1} and \texttt{PhysioNet 0.5}, respectively.
For the forecast task, we validated our method on three datasets with varying feature dimensions, which are collected and preprocessed in GluonTS~\citet{alexandrov2020gluonts}. We adopt root mean square error (RMSE), mean absolute error (MAE) and continuous ranked probability score (CRPS) for evaluation.
Details can be found in Appendix 2.1.

\paragraph{Baselines}
We focused on comparing CSBI++ with its vanilla version (CSBI) on time series modeling tasks.
For the imputation task, our baselines additionally include V-RIN~\citep{mulyadi2021vrin}, multitask Gaussian process (multitask GP)~\citep{durichen2014multitaskgp} and GP-VAE~\citep{fortuin2020gpvae}.
For the forecasting task, our baselines additionally include GP-copula~\citep{salinas2019gpcopula}, Vec-LSTM-low-rank-Coupula (VEC)~\citep{salinas2019gpcopula} and TransMAF~\citep{rasul2021transmaf}.

\paragraph{Results}
For the imputation task, \Cref{tab: exp_time} \textbf{(Left)} illustrates that CSBI++ surpasses CSBI in all evaluated metrics across the three datasets, demonstrating the effectiveness of integrating Hutch++ into an SB-based model for multidimensional data modeling.
For the forecasting task, \Cref{tab: exp_time} \textbf{(Right)} shows that CSBI++ demonstrates superior performance relative to its standard version and exhibits substantial enhancements on \texttt{Solar}. It is noteworthy that the improvement on electricity with a dimension of 370 further underscores the effectiveness and applicability of our method in modeling high-dimensional data.
We provide examples in \Cref{fig: time} and more visualization in Appendix 3.2.
\begin{figure}[!t]
    \centering
    \begin{subfigure}{0.47\linewidth}
        \centering
        \includegraphics[width=\linewidth]{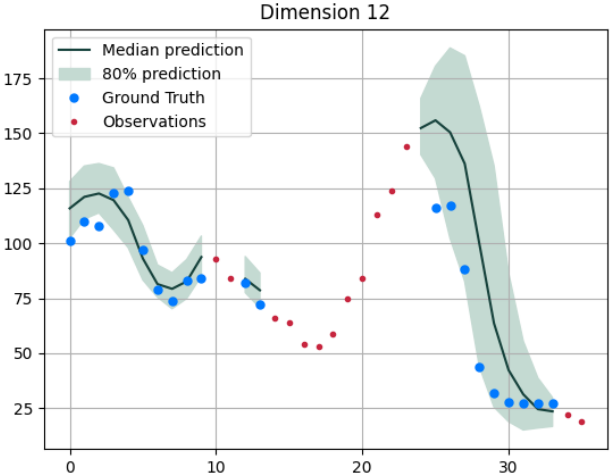}
        \caption{}
        \label{fig: time_i}
    \end{subfigure}
    \begin{subfigure}{0.47\linewidth}
        \centering
        \includegraphics[width=\linewidth]{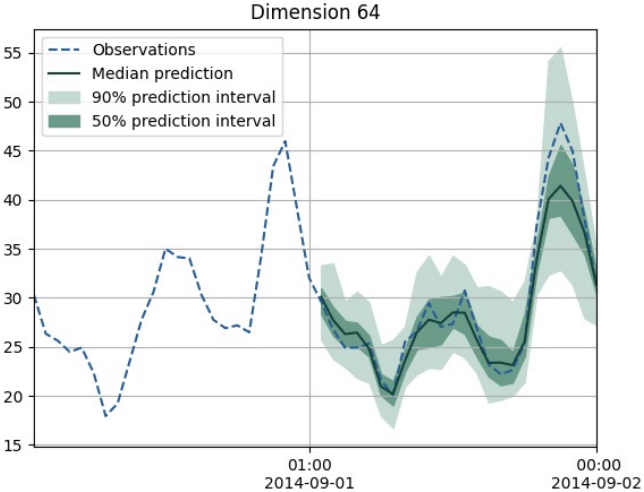}
        \caption{}
        \label{fig: time_f}
    \end{subfigure}
    \vspace{-1mm}
    \caption{\small \textbf{(a)Imputation} examples for \texttt{PM25} for 1 (out of 36) dimensions. \textbf{(b)Forecasting} examples for \texttt{Electricity} for 1 (out of 370) dimensions}
    \vspace{-2mm}
    \label{fig: time}
\end{figure}

\subsection{Image Data Modeling}\label{sec: image}
In light of the demonstrated effectiveness of our proposed method in simulations and its encouraging performance in multidimensional time series modeling, we further investigated its potential to handle high-dimensional data.
We apply Hutch++ to advanced SB-based diffusion models and compare their performance in image data modeling.

\paragraph{Experiment Setup}
We first proposed SB-FBSDE++ based on SB-FBSDE~\citep{forward_backward_SDE} which is a very popular SB-based generative model.
We compare their performance on \texttt{MNIST}, \texttt{Fasion-MNIST}, and \texttt{CIFAR10} in terms of negative log-likelihood. 
Moreover, to better validate our method in modeling images with complex patterns, we employed VSDM~\citep{VSDM} as the base model, a state-of-the-art SB-based diffusion model which is capable of tuning-friendly training on large-scale data.
We denote our improved model incorporating the Hutch++ estimator as VSDM++.

\paragraph{Results}
~\Cref{tab: image} \textbf{(Top)} shows the comparison of the NLL values evaluated on the test set. 
SB-FBSDE++ achieves consistent improvements across all three datasets.
Especially on the \texttt{Fashion-MNIST} dataset, SB-FBSDE++ shows promising improvements, suggesting that our proposed method can effectively adapt to SB-based diffusion models for modeling images with complex patterns.
Furthermore, we compared the FID scores recorded in ~\Cref{tab: image} \textbf{(Bottom)} throughout the training process on \texttt{CIFAR10} datasets. 
The improvements introduced by VSDM++ throughout the training process demonstrate that our approach remains effective on high-dimensional data, highlighting its potential for more complex generative modeling.
The generated images are provided in Appendix 3.3.

\begin{table}[!t]
    \centering
    \caption{\small \textbf{(Top)} Image data modeling evaluation using negative log-likelihood (NLL; bits/dim) on the test set. \textbf{(Bottom)} Comparison of convergence speed of FID scores for VSDM and VSDM++.}
        \label{tab: image}
    \begin{minipage}{0.48\textwidth} 
        \centering
        \scalebox{0.75}{
        \begin{tabular}{lccc}
        \toprule[1.5pt]
        Method &MNIST &FASION-MNIST &CIFAR10\\
        \midrule
        SB-FBSDE* &0.95 &1.63 &2.96\\ 
        \textbf{SB-FBSDE++ (Ours)} &\textbf{0.91} &\textbf{1.56} & \textbf{2.93}\\
        \bottomrule
        \end{tabular}}
    \end{minipage}
    \begin{minipage}{0.48\textwidth} 
    \vspace{3mm}
        \centering
        \scalebox{0.75}{
        \begin{tabular}{lcccccccc}
        \toprule[1.5pt]
        K Images &10K &30K &50K & 100K & 150K & 200K \\
        \midrule
        VSDM* &\textbf{13.97} &7.12 &5.56 &3.64 &3.27 &3.04 \\ 
        \textbf{VSDM++ (Ours)} &14.28 &\textbf{6.93} &\textbf{5.31} &\textbf{3.24} &\textbf{3.15} &\textbf{2.97} \\
        \bottomrule
        \end{tabular}}
    \end{minipage}
\end{table}

\section{Conclusions}

Divergence estimation has been a critical challenge for diffusion models with optimal transport properties. However, the issue of large variance significantly impacts scalability concerning data size and dimension, making these models less effective for large-scale real-world generation tasks. To address this, we explore the use of Hutch++ to minimize variance in trace estimation within optimal transport-based generative models, observing significant speedups in training diffusion models while preserving optimal transportation plans. Our work paves the way for optimizing transport maps and shows potential for a deeper understanding of optimal transport in various scientific studies.



\clearpage
\newpage
\bibliography{aistats2025}
\bibliographystyle{apalike}

\section*{Checklist}


 \begin{enumerate}

 \item For all models and algorithms presented, check if you include:
 \begin{enumerate}
   \item A clear description of the mathematical setting, assumptions, algorithm, and/or model. [Yes]
   \item An analysis of the properties and complexity (time, space, sample size) of any algorithm. [Yes]
   \item (Optional) Anonymized source code, with specification of all dependencies, including external libraries. [Yes]
 \end{enumerate}

 \item For any theoretical claim, check if you include:
 \begin{enumerate}
   \item Statements of the full set of assumptions of all theoretical results. [Yes]
   \item Complete proofs of all theoretical results. [Yes]
   \item Clear explanations of any assumptions. [Yes]     
 \end{enumerate}

 \item For all figures and tables that present empirical results, check if you include:
 \begin{enumerate}
   \item The code, data, and instructions needed to reproduce the main experimental results (either in the supplemental material or as a URL). [Yes]
   \item All the training details (e.g., data splits, hyperparameters, how they were chosen). [Yes]
         \item A clear definition of the specific measure or statistics and error bars (e.g., with respect to the random seed after running experiments multiple times). [Yes]
         \item A description of the computing infrastructure used. (e.g., type of GPUs, internal cluster, or cloud provider). [Yes]
 \end{enumerate}

 \item If you are using existing assets (e.g., code, data, models) or curating/releasing new assets, check if you include:
 \begin{enumerate}
   \item Citations of the creator If your work uses existing assets. [Yes]
   \item The license information of the assets, if applicable. [Not Applicable]
   \item New assets either in the supplemental material or as a URL, if applicable. [Yes]
   \item Information about consent from data providers/curators. [Not Applicable]
   \item Discussion of sensible content if applicable, e.g., personally identifiable information or offensive content. [Not Applicable]
 \end{enumerate}

 \item If you used crowdsourcing or conducted research with human subjects, check if you include:
 \begin{enumerate}
   \item The full text of instructions given to participants and screenshots. [Not Applicable]
   \item Descriptions of potential participant risks, with links to Institutional Review Board (IRB) approvals if applicable. [Not Applicable]
   \item The estimated hourly wage paid to participants and the total amount spent on participant compensation. [Not Applicable]
 \end{enumerate}

 \end{enumerate}

\clearpage
\onecolumn
\onecolumn
\aistatstitle{Supplementary Materials for ``Optimal Stochastic Trace Estimation in Generative Modeling''}


\vspace{-5mm}
\setcounter{section}{0}
\section{Proofs for Theoretical Analysis}\label{sec:appendix_trace_estimate}
\subsection{Proof of Lemma 5.1}
The following result gives us an estimate of the expectation and variance for the Hutchinson trace estimator:\\
To compute the expectation:
\begin{align*}
  \E[H_m(\bA)] &= \frac{1}{m}\sum_{i=1}^{m}\E[\mathbf{v}_i^\top\bA\mathbf{v}_i] \\
  &= \E[\mathbf{v}_1^\top \bA \mathbf{v}_1] \\
  &= \sum_{i,j=1}^{n}\E[\bA_{ij}\mathbf{v}_{1,i}\mathbf{v}_{1,j}] \\
  &= \sum_{i, j=1}^{n}\bA_{ij}\delta_{ij} = \sum_{i=1}^{n}\bA_{ii} = \mathrm{Tr}(\bA).
\end{align*}
For the variance, let $\bA=\bO^\top\bL\bO$ be the eigendecomposition of $\bA$:
\begin{align*}
  \Var[H_m(\bA)] &= \frac{1}{m}\Var[\mathbf{v}_1^\top \bA \mathbf{v}_1] \\
  &= \frac{1}{m}\Var[\mathbf{v}_1^\top\bO^\top \bL \bO \mathbf{v}_1] \\
  &= \frac{1}{m}\Var\left[ (\bO\mathbf{v}_1)^\top \bL (\bO\mathbf{v}_1) \right] \\
  &= \frac{1}{m}\Var\left[ \sum_{i=1}^{D}\lambda_i |\mathbf{w}_{1,i}|^2 \right] \qquad (\text{$\mathbf{w}_1:=\bO\mathbf{v}_1$})\\
  &= \frac{1}{m}\sum_{i=1}^{D}\lambda_i^2 \Var[|\mathbf{w}_{1, i}|^2] = \frac{2}{m}\|\bA\|_F^2\le \frac{m}{2}\mathrm{Tr}^2(\bA).
\end{align*}

\subsection{Analysis of Hutch++ estimator}
\label{sec:appendix_hutchpp}
The key observation here is that $\|\mathbf{A}\|_F \approx \mathrm{Tr}(\mathbf{A})$ holds when $\mathbf{A}$ has only a few large eigenvalues. This happens because in such cases, the Frobenius norm, which sums the squares of the eigenvalues, is dominated by the largest ones, making it comparable to the trace, which sums the eigenvalues. 

On the other hand, for matrices with many small eigenvalues, we observe a different behavior. In such cases, the Frobenius norm and the trace diverge. Specifically, when most eigenvalues are small, we have:
\[
\|\mathbf{A}\|_F \approx \sqrt{n} \lambda_1 = \frac{1}{\sqrt{n}} n \lambda_1 \approx \frac{1}{\sqrt{n}} \mathrm{Tr}(\mathbf{A}),
\]
where $\lambda_1$ is the largest eigenvalue and $n$ is the matrix dimension. This means the Frobenius norm grows faster than the trace when there are many small eigenvalues.
Taking advantage of this dichotomy, Meyer et al. proposed the Hutch++ algorithm, which cleverly combines a low-rank approximation for large eigenvalues with Hutchinson’s estimator for the smaller ones. The algorithm is as follows:
$$
H_m^{++}(\mathbf{A}) = \mathrm{Tr}(Q^\top\mathbf{A}Q) + H_{\frac{m}{3}}((I-QQ^\top)\mathbf{A}),
$$
\newpage
where $Q^\top\mathbf{A}Q$ is a good low-rank approximation of $\mathbf{A}$ that captures the large eigenvalues, while the second term uses the Hutchinson estimator to approximate the trace of the remaining smaller eigenvalues.

\begin{lemma}
Let $\mathbf{A}$ be an $n \times n$ PSD (positive semidefinite) matrix, and let $\mathbf{A}_k$ denote the best rank-$k$ approximation of $\mathbf{A}$. Then,
\[
\|\mathbf{A} - \mathbf{A}_k\|_F \le \frac{1}{2\sqrt{k}} \mathrm{Tr}(\mathbf{A}).
\]
\end{lemma}

We begin by expressing the Frobenius norm of the difference between $\mathbf{A}$ and its rank-$k$ approximation in terms of the eigenvalues:
\[
\frac{\|\mathbf{A} - \mathbf{A}_k\|_F}{\mathrm{Tr}(\mathbf{A})} = \frac{\sqrt{\sum_{i=k+1}^{n} \lambda_i^2}}{\sum_{i=1}^{n} \lambda_i},
\]
where $\lambda_i$ are the eigenvalues of $\mathbf{A}$.

Using an inequality that approximates this expression, we can write:
\[
\frac{\|\mathbf{A} - \mathbf{A}_k\|_F}{\mathrm{Tr}(\mathbf{A})} \le \frac{\sqrt{\lambda_{k+1} \sum_{i=k+1}^{n} \lambda_i}}{\sum_{i=1}^{k} \lambda_i + \sum_{i=k+1}^{n} \lambda_i}.
\]
Let $a = \lambda_{k+1}$, $b = \sum_{i=1}^{k} \lambda_i$, and $\lambda = \sum_{i=k+1}^{n} \lambda_i$. The expression simplifies to
\[
\frac{\|\mathbf{A} - \mathbf{A}_k\|_F}{\mathrm{Tr}(\mathbf{A})} \le \frac{\sqrt{a\lambda}}{b + \lambda}.
\]
To minimize this expression, let $f(\lambda) = \frac{\sqrt{a\lambda}}{b + \lambda}$. We take the derivative:
\[
f'(\lambda) = \frac{a(b - \lambda)}{2\sqrt{a\lambda}(b + \lambda)^2}.
\]
Setting $f'(\lambda) = 0$, we find that the critical point occurs at $\lambda = b$, where the function attains its maximum.

Substituting $\lambda = b$ into the expression for $f(\lambda)$ gives:
\[
\frac{\|\mathbf{A} - \mathbf{A}_k\|_F}{\mathrm{Tr}(\mathbf{A})} \le f(b) = \frac{1}{2} \sqrt{\frac{a}{b}} = \frac{1}{2} \sqrt{\frac{\lambda_{k+1}}{\sum_{i=1}^{k} \lambda_i}} \le \frac{1}{2} \sqrt{\frac{\lambda_{k+1}}{k \lambda_{k+1}}} = \frac{1}{2\sqrt{k}}.
\]
Thus, we have the desired result.

For low-rank approximations, one can use a randomized approach by drawing a random $D \times (k + p)$ matrix $S$.

\begin{theorem}[\cite{halko2011finding}]
Let $Q$ be any orthonormal basis for $\mathbf{A} S$. Then,
\begin{equation}
    \mathbb{E}\left[ \|(I-QQ^\top) \mathbf{A}\|_F^2 \right] \le \left( 1 + \frac{k}{p-1} \right) \|\mathbf{A} - \mathbf{A}_{k}\|_F^2.
\end{equation}
\end{theorem}

\subsection{Proof of Lemma 5.2}
For the expectation of Hutch++, by law of total expectation $\E[\E[X|Y]=\E[X]$, we can show
\begin{align}
    \E\left[ H_m^{++}(\bA) \right]&=\E\left[ \mathrm{Tr}(Q^\top \bA Q)+H_{\frac{m}{3}}( (I-QQ^\top)\bA) \right]\\
    &=\E\left[ \mathrm{Tr}(Q^\top \bA Q) \right]+\E\left[ \E\left[ H_{\frac{m}{3}}( (I-QQ^\top)\bA)|Q \right] \right]\\
    &=\E\left[ \mathrm{Tr}(Q^\top \bA Q) \right]+\E\left[ \mathrm{Tr}\left( (I-QQ^\top)\bA \right) \right]\\
    &=\E[\mathrm{Tr}(QQ^\top\bA)]+\E\left[ \mathrm{Tr}(\bA-QQ^\top\bA) \right]\\
    &=\E[\mathrm{Tr}(\bA)]=\mathrm{Tr}(\bA). 
\end{align}

Using the conditional variance formula for random variables $X$ and $Y$:
\[
\Var[X] = \E[\Var[X|Y]] + \Var(\E[X|Y]),
\]
we let $X = H_m^{++}(\bA)$ and $Y = Q$ to get
\begin{align}
    &\E\left[ \Var\left[ H_m^{++}(\bA)|Q \right] \right]\\
    &=\E\left[ \Var\left[ \mathrm{Tr}(Q^\top \bA Q)+H_{\frac{m}{3}}\left( (I-QQ^\top)\bA \right)|Q \right] \right]\\
    &=\E\left[ \Var\left[ H_{\frac{m}{3}}\left( (I-QQ^\top)\bA\right)|Q \right]\right]\\
    &\le \frac{6}{m}\E\left[ \|(I-QQ^\top)\bA\|_{F}^2 \right]\\
    &\le \frac{6}{m}\left( 1+\frac{k}{p-1} \right)\|\bA-\bA_{k}\|_{F}^2,
\end{align}
while
\begin{align}
    &\Var\left[ \E\left[ H_m^{++}(\bA)|Q \right] \right]\\
    &=\Var\left[ \E\left[ \mathrm{Tr}(Q^\top \bA Q)+H_{\frac{m}{3}}( (I-QQ^T)\bA)|Q \right] \right]\\
    &=\Var\left[ \mathrm{Tr}(Q^\top \bA Q)+\mathrm{Tr}( (I-QQ^\top)\bA) \right]\\
    &=\Var[\mathrm{Tr}(\bA)]=0.
\end{align}
Now we choose $p$, $k$ such that $p=k+1$ and $k+p=\frac{m}{3}$, i.e., $p=\frac{m+3}{6}, k=\frac{m-3}{6}$ we get that 
$$\Var[H_m^{++}(\bA)]\le \frac{12}{m}\|\bA-\bA_k\|_F^2\le \frac{12}{m}\frac{1}{4k}\mathrm{Tr}^2(\bA)=\frac{3}{km}\mathrm{Tr}^2(\bA)=\frac{18}{m(m-3)}\mathrm{Tr}^2(\bA).$$
For i.i.d. sub-Gaussian vairables, we have the following Hanson-Wright inequality
\begin{lemma} Let $\mathbf{v}\in\R^n$ be a vector of a vector of i.i.d. sub-Gaussian random variables of mean zero. Then it holds
  $$\PP\left( |\mathbf{v}^\top \bar\bA\mathbf{v}-\mathbb{E}[\mathbf{v}^\top \bar\bA\mathbf{v}]|>t \right)\le 2\exp\left( -c \cdot \min \left\{ \frac{t^2}{\|\bar{\mathbf{A}}\|_F^2}, \frac{t}{\|\bar{\bA}\|_2} \right\} \right).$$
  \label{lemma:HansonWright}
  Here $\|\bar{\bA}\|_2=\max_{\mathbf{w}\in\R^n}\|\bar\bA\mathbf{w}\|_2/\|\mathbf{w}\|_2$. 
\end{lemma}
Applying Lemma \ref{lemma:HansonWright} to $\bar{\bA}=\mathrm{diag}\{\bA, \cdots, \bA\}$, we can show
Following a similar argument in \cite{hutch_pp}, it holds that 
\begin{equation}
\begin{aligned}
  \PP(\varepsilon(H_m)\ge \varepsilon)= \PP\left( \frac{|H_m(\bA)-\mathrm{Tr}(\bA)|}{\mathrm{Tr}(\bA)}\ge \varepsilon \right)\le\delta\Rightarrow C\sqrt{\frac{\log(1/\delta)}{m}}<\varepsilon\Rightarrow m=\mathcal{O}\left(\frac{\log(1/\delta)}{\varepsilon^2}  \right).
\end{aligned}
\label{}
\end{equation}
With the quadratic improvement for the variance estimate, we obtain that 
  \begin{equation}
	\begin{aligned}
		\PP(\varepsilon(H_m^{++})\ge \varepsilon)= \PP\left( \frac{|H^{++}_m(\bA)-\mathrm{Tr}(\bA)|}{\mathrm{Tr}(\bA)}\ge \varepsilon \right)\le \delta\Rightarrow m=\mathcal{O}\left(\frac{\log(1/\delta)}{\varepsilon^2}+\log(1/\delta)  \right).
	\end{aligned}
	\label{}
\end{equation}

\subsection{Proof of Proposition 5.4}
Again, we use the law of total expectation to show
\begin{align*}
    \E\left[ \tilde{H}_m^{++}(\tilde{\bA}) \right]&=\E\left[ \mathrm{Tr}(Q^\top \tilde{\bA} Q)+H_{\frac{m}{3}}( (I-QQ^\top)\tilde{\bA}) \right]\\
    &=\E\left[ \mathrm{Tr}(Q^\top \tilde{\bA} Q) \right]+\E\left[ \E\left[ H_{\frac{m}{3}}( (I-QQ^\top)\tilde{\bA})|Q \right] \right]\\
    &=\E\left[ \mathrm{Tr}(Q^\top \tilde{\bA} Q) \right]+\E\left[ \mathrm{Tr}\left( (I-QQ^\top)\tilde{\bA} \right) \right]\\
    &=\E[\mathrm{Tr}(QQ^\top\tilde{\bA})]+\E\left[ \mathrm{Tr}(\tilde{\bA}-QQ^\top\tilde{\bA}) \right]\\
    &=\E[\mathrm{Tr}(\tilde{\bA})]=\mathrm{Tr}(\tilde{\bA}). 
  \end{align*}
Now we apply the conditional variance formula for $X = \tilde{H}_m^{++}(\tilde{\bA})$ and $Y = Q$
\begin{align*}
	&\E\left[ \Var\left[ \tilde{H}_m^{++}(\tilde{\bA})|Q \right] \right]\\
	&=\E\left[ \Var\left[ \mathrm{Tr}(Q^\top \tilde{\bA} Q)+H_{\frac{m}{3}}\left( (I-QQ^\top)\tilde{\bA} \right)|Q \right] \right]\\
	&=\E\left[ \Var\left[ H_{\frac{m}{3}}\left( (I-QQ^\top)\tilde{\bA}\right)|Q \right]\right]\\
	&\le \frac{6}{m}\E\left[ \|(I-QQ^\top)\tilde{\bA}\|_{F}^2 \right]\\
    &\le \frac{12}{m}\E\left[ \|(I-QQ^\top)\bA\|_{F}^2 \right]+\frac{12}{m}  \E\left[\|(I-QQ)^\top(\tilde{\bA}-\bA) \|^2_F \right]              \\
	&\le \frac{12}{m}\left[\left( 1+\frac{k}{p-1} \right)\|\bA-\bA_{k}\|_{F}^2+\mathcal{O}((L_s-1)^2\eta^2)\right]\\
	&\le\frac{36}{m(m-3)}\mathrm{Tr}^2(\bA)+\frac{1}{m}\mathcal{O}((L_s-1)^2\eta^2),
\end{align*}
where we choose $p=\frac{m+3}{6}, k=\frac{m-3}{6}$. Meanwhile, 
\begin{align*}
	&\Var\left[ \E\left[ \tilde{H}_m^{++}(\tilde{\bA})|Q \right] \right]\\
	&=\Var\left[ \E\left[ \mathrm{Tr}(Q^\top \tilde{\bA} Q)+H_{\frac{m}{3}}( (I-QQ^T)\tilde{\bA})|Q \right] \right]\\
	&=\Var\left[ \mathrm{Tr}(Q^\top \tilde{\bA} Q)+\mathrm{Tr}( (I-QQ^\top)\tilde{\bA}) \right]\\
	&=\Var[\mathrm{Tr}(\tilde{\bA})]=0.
\end{align*}
then we could get that 
$$\mathrm{Var}[\tilde{H}^{++}(\tilde{\bA})]\le \frac{36}{m(m-3)}\mathrm{Tr}^2(\bA)+\frac{1}{m}\mathcal{O}((L_s-1)^2\eta^2).$$

\subsection{Complexity analysis}
The computational complexity of Hutch++ for $\bA\in \R^{D\times D}$ mainly comes from two parts:

Part I: multiplying $\bA$ with a thin matrix $Q$ requires $O(D^2m)$ operations;

Part II: conducting QR decomposition for the span of $\bA \bS$  to get an orthogonal matrix $Q\in \mathbb{R}^{D\times \mathcal{O}(m)}$ requires $\mathcal{O}(c D m^2)$, where $c > 1$ denotes the slowdown factor in computations of matrix decomposition compared to matrix multiplication due to the parallelism and hardware optimization in GPUs.

Specifically, our Hutch++ proposes conducting QR decomposition every $L_s$ iterations, which leads to a reduced computation $
\mathcal{O}\left(\frac{c D m^2}{L_s}\right)$ in part II on average. By Proposition 5.4, our approximate
Hutch++ needs
\begin{equation} \mathcal{O}\bigg(\frac{c D m^2}{L_s} + D^2 m\bigg) \text{ operations to achieve a variance } \mathcal{O}\bigg(\frac{1}{m^2} + \frac{\sigma^2}{m}\bigg) \end{equation}

where $\sigma^2=\mathcal{O}((L_s-1)^2 \eta^2)$.

Meanwhile,  the vanilla Hutchinson estimator requires $\mathcal{O}(m D^2)$ operations to achieve a variance of $\mathcal{O}\left(\frac{1}{m}\right)$. To achieve a similar scale of variance as our approximate Hutch++, Hutch requires a complexity $$\mathcal{O}\bigg(\frac{D^2}{\frac{1}{m^2} + \frac{\sigma^2}{m}}\bigg)=\mathcal{O}\bigg(\frac{D^2 m^2}{1 + m\sigma^2}\bigg)$$

Given smooth enough changes of the divergence $A_t$ w.r.t. $t$ such that $\sigma<1$, we can see Hutch++ is cheaper when

\begin{equation} \begin{aligned} & \underbrace{\mathcal{O}\bigg(\frac{c Dm^2}{L_s}+D^2m\bigg)}_{\text{Approximate Hutch++ cost}}<\underbrace{\mathcal{O}\bigg(\frac{D^2 m^2}{1+m\sigma^2}\bigg)}_{\text{Vinilla Hutch cost}}. \end{aligned}\label{eqn:costcomp} \end{equation}


For $c\gg 1$ and $\sigma<1$, we can choose $ L_s\approx \frac{cm}{D}$ such that (\ref{eqn:costcomp}) is valid, which demonstrates the necessity of reusing eigenvectors along the trajectory of the integral for Hutch++.

Similarly, instead of reusing eigenvectors only at neighboring points within the same iteration, they can also be reused across subsequent iterations, with a comparable analysis applying. Strategies that reuse previous gradients every \(K\) iterations have achieved computational success, including Stochastic Variance Reduced Gradient (SVRG) \cite{johnson2013accelerating} and Federated Averaging (FedAvg) \cite{mcmahan2017communication, li2019convergence}.

\subsection{Proof of Proposition 5.5}
In Neural ODE, the log-likelyhood could be computed by 
\begin{equation}
  \log p_\theta (\mathbf{z}(t_1))=\log q(\mathbf{z}(t_0))-\int_{t_0}^{t_1}\mathrm{Tr}\left( \frac{\pa f}{\pa \mathbf{z}} \right)dt.
  \label{}
\end{equation}
After applying the approximate Hutch++ estimator, we have 
\begin{equation}
  \log \tilde{p}_\theta(t_1)=\log q(\mathbf{z}(t_0))-\int_{t_0}^{t_1}\tilde{H}^{++}_m\left( \frac{\pa f}{\pa \mathbf{z}} \right)dt.
  \label{}
\end{equation}
Taking the expectation and use the fact that $\E[\tilde{H}^{++}_m(\bA)]=\mathrm{Tr}(\bA)$, we get that 
\begin{align*}
  \E[\log \tilde{p}_\theta(t_1)]&=\log q(\mathbf{z}(t_0))-\E\int_{t_0}^{t_1}\tilde{H}^{++}_m\left( \frac{\pa f}{\pa \mathbf{z}} \right)dt
  \\
  &=\log q(\mathbf{z}(t_0))-\int_{t_0}^{t_1}\E\left[ \tilde{H}^{++}_m\left( \frac{\pa f}{\pa \mathbf{z}} \right) \right]dt\\
  &=\log q(\mathbf{z}(t_0))-\int_{t_0}^{t_1} \mathrm{Tr}\left( \frac{\pa f}{\pa \mathbf{z}} \right)dt\\
  &=\log p_\theta(\mathbf{z}(t_1)).
\end{align*}
For the variance estimate:
\begin{align*}
  \mathrm{Var}[\log \tilde{p}_\theta(t_1)]&=\E|\log \tilde{p}_\theta(t_1)-\log p_\theta(\mathbf{z}(t_1))|^2\\
  &=\E\left|\int_{t_0}^{t_1}\left[\tilde{H}^{++}_m\left( \frac{\pa f}{\pa \mathbf{z}} \right)-\mathrm{Tr}\left( \frac{\pa f}{\pa \mathbf{z}} \right)\right]dt\right|^2\\
  &\le (t_1-t_0)\E\left[\int_{t_0}^{t_1}\left|\tilde{H}^{++}_m\left( \frac{\pa f}{\pa \mathbf{z}} \right)-\mathrm{Tr}\left( \frac{\pa f}{\pa \mathbf{z}} \right)\right|^2 dt\right]\\
  &=(t_1-t_0)\int_{t_0}^{t_1}\E\left[ \tilde{H}^{++}_m\left( \frac{\pa f}{\pa \mathbf{z}} \right)-\mathrm{Tr}\left( \frac{\pa f}{\pa \mathbf{z}} \right) \right]^2 dt\\
  &= (t_1-t_0)\int_{t_0}^{t_1}\mathrm{Var}\left[ \tilde{H}^{++}_m\left( \frac{\pa f}{\pa \mathbf{z}} \right) \right]dt\\
  &\le (t_1-t_0)\int_{t_0}^{t_1}\left[\frac{36}{m(m-3)}M^2+\frac{1}{m}\mathcal{O}((L_s-1)^2\eta^2)\right]dt\\
  &\le (t_1-t_0)^2\left[ \frac{36}{m(m-3)}M^2+\frac{1}{m}\mathcal{O}((L_s-1)^2\eta^2) \right].
\end{align*}

\section{Experimental Details}
\subsection{Datasets}

\paragraph{Simulation data}
We compare the FFJORD++ with FFJORD on several toy distributions, including \texttt{2spirals}, \texttt{checkerboard}, \texttt{rings} and \texttt{circles}.

\paragraph{Time series dataset for imputation }
We run experiments for time series imputation on two datasets. 
\texttt{PM2.5}, a air quality dataset~\citep{yi2016pm25}, consists of the hourly sampled PM2.5 air quality index from 36 monitoring stations for 12 months. Approximately 13\% of the data is missing, and the patterns of missing values are not random. The ground truth values at the missing points are known and we use them as the test data set.
\texttt{PhysioNet}, a healthcare dataset in PhysioNet Challenge 2012~\citep{silva2012physio}, consists of 4000 clinical time series with 35 variables for 48 hours from intensive care unit (ICU). Following the processing method in previous work~\citep{provably_schrodinger_bridge, CSDI, cao2018data_precoss1, che2018data_precoss2}, we aggregate data to one point per hour. The missing rate of the dataset is about 80\%. We randomly mask a specific percentage (10\%, 50\%) of the data that comprise the test dataset, denoted as \texttt{PhysioNet 0.1} and \texttt{PhysioNet 0.5}, respectively.

\paragraph{Time series dataset for forecasting}
For the forecast task, we validated our method on three datasets with varying feature dimensions, which are collected and preprocessed in GluonTS~\citet{alexandrov2020gluonts}.
\Cref{tab: time_stat} summarizes the properties of these datasets including the feature dimension, total time steps, history steps and prediction steps for each datasets.

\begin{table}[!h]
    \centering
    \scalebox{0.9}{
    \begin{tabular}{c|c|c|c|c}
    \toprule
     \multirow{2}{*}{} & feature &total &history &prediction\\ 
     & dimension & time step & steps & steps \\
     \midrule
     Exchange & 8& 6071& 180& 30\\
     Solar & 137& 7009& 168& 24\\
     Electricity & 370 &5833& 168& 24\\
    \bottomrule
    \end{tabular}}
    \caption{\small Overview of the datasets used in time series forecasting}
    \label{tab: time_stat}
\end{table}

\subsection{Implementation Details}
We set the same hyper-parameters for each pair of both base models and our improved models which employ Hutch++ estimator.
Specifically, we set the learning rate to 5e-4 and use the second-order midpoint solver as the default ODE solver, with the same number of function evaluations of 100 in the simulation.
Additionally, we used all the parameters from their corresponding open-source code for the remaining experiments, without making any modifications.
As discussed in Section 6.1, we set the default shared step size $L_s = 10$ for all experiments. For models training with random time point in SB-based diffusion models, we re-design the training strategy based on continuous time interval and update $Q$ every $L_s$ steps.

\subsection{Computing Resources}
For all experiments, we utilized the PyTorch \citep{paszke2019pytorch} framework to implement all methods and trained models with NVIDIA GeForce RTX 4090 and RTX A6000 GPUs.

\subsection{Limitations}
To avoid computing $\frac{\partial z_t}{\partial \mathbf{x}(t)} S$ directly, 
the Jcacobian-vector products  $\frac{\partial z_t}{\partial \mathbf{x}(t)} S$ can be computed utilizing advanced deep learning frameworks through two rounds of reverse-mode automatic differentiation, whose time cost can be neglected compared to forward and backward propagation in the neural network.
However, the gradients collected in the reverse-mode automatic differentiation lead to large and non-negligible memory costs.
Therefore, how to more efficiently train divergence-based likelihood in diffusion models with optimal guarantee is still the promising direction in future work.

\section{Additional Experimental Results}
\subsection{Simulations}
\begin{figure}[!t]
    \centering
    \begin{subfigure}[b]{0.48\textwidth}
        \centering
        \includegraphics[width=\textwidth]{fig/2s_convergence.pdf}
        \caption{2spirals}
        \label{fig: 2s}
    \end{subfigure}
    \hfill
    \begin{subfigure}[b]{0.48\textwidth}
        \centering
        \includegraphics[width=\textwidth]{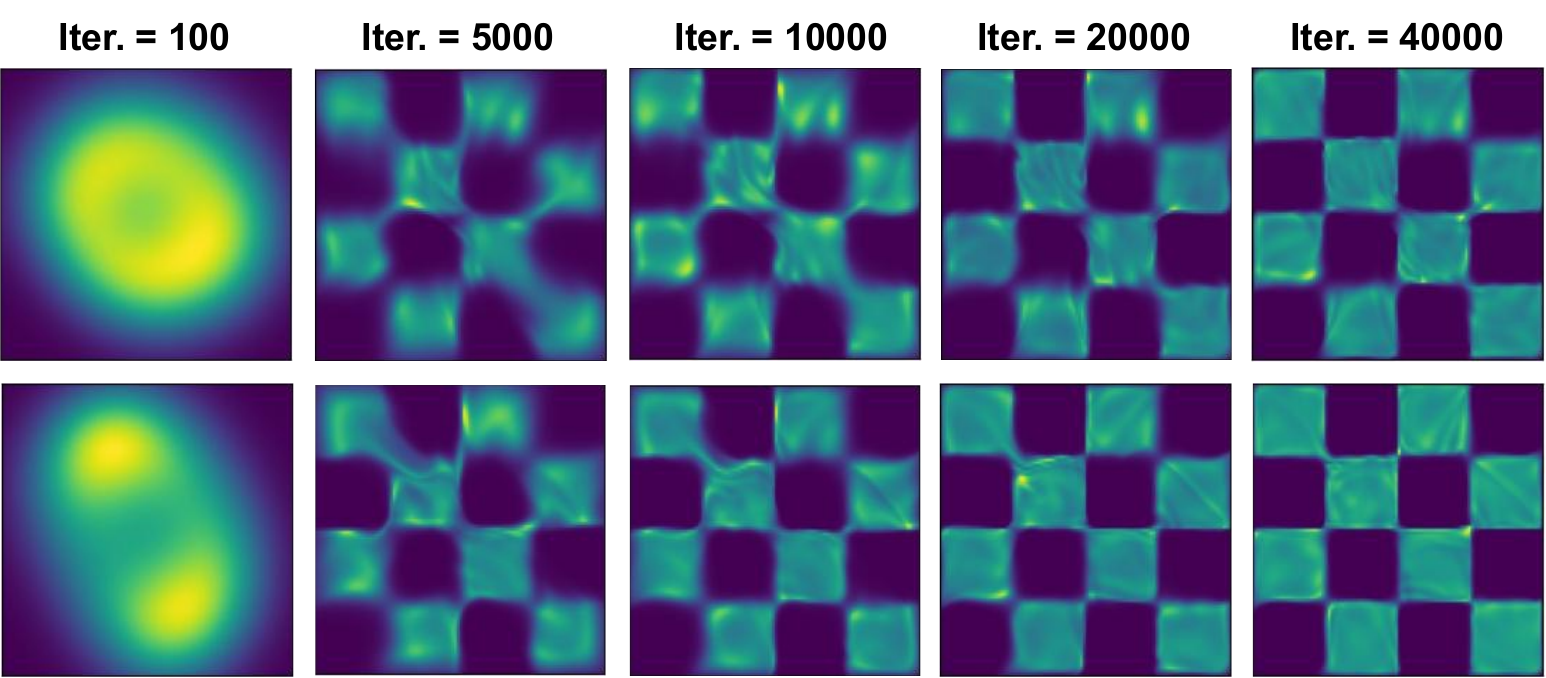}
        \caption{checkerboard}
        \label{fig: cb}
    \end{subfigure}
    \begin{subfigure}[b]{0.48\textwidth}
        \centering
        \includegraphics[width=\textwidth]{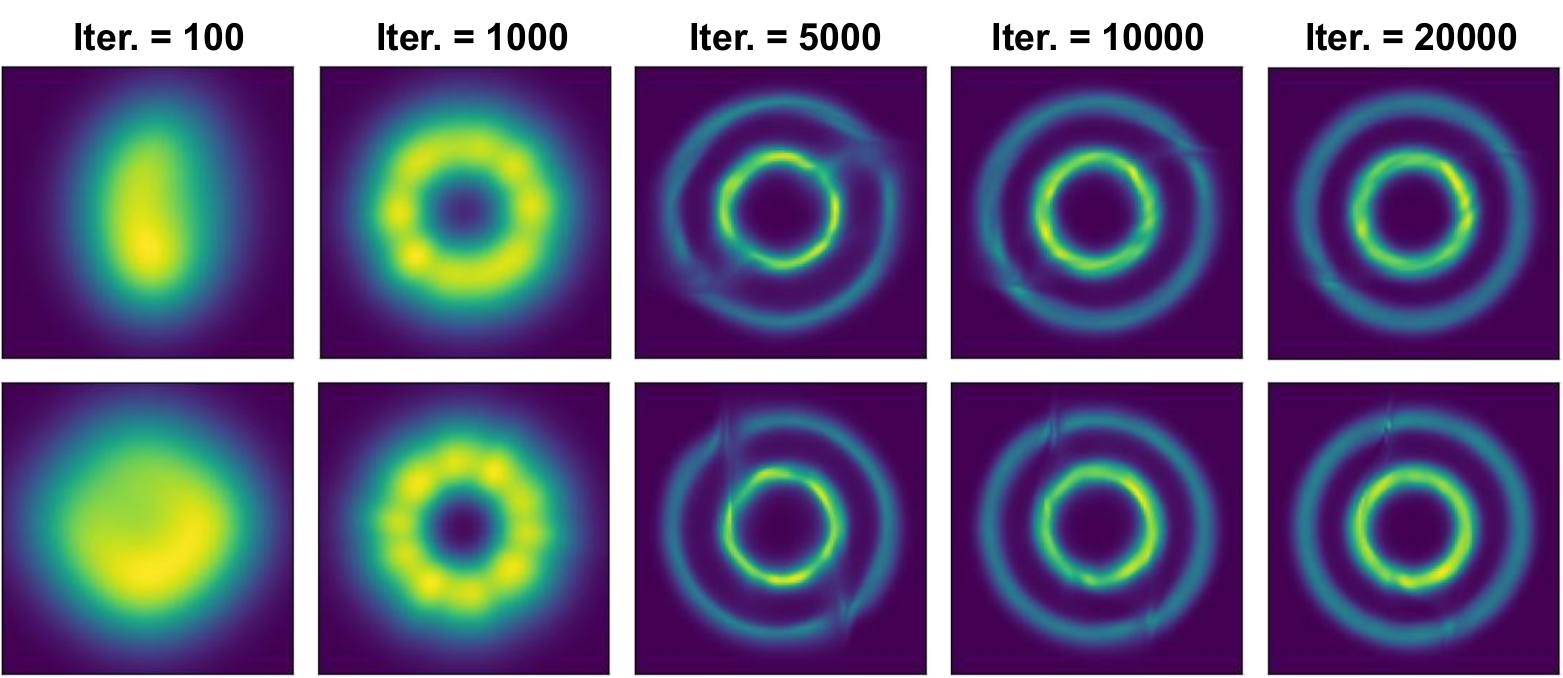}
        \caption{rings}
        \label{fig: 2s}
    \end{subfigure}
    \hfill
    \begin{subfigure}[b]{0.48\textwidth}
        \centering
        \includegraphics[width=\textwidth]{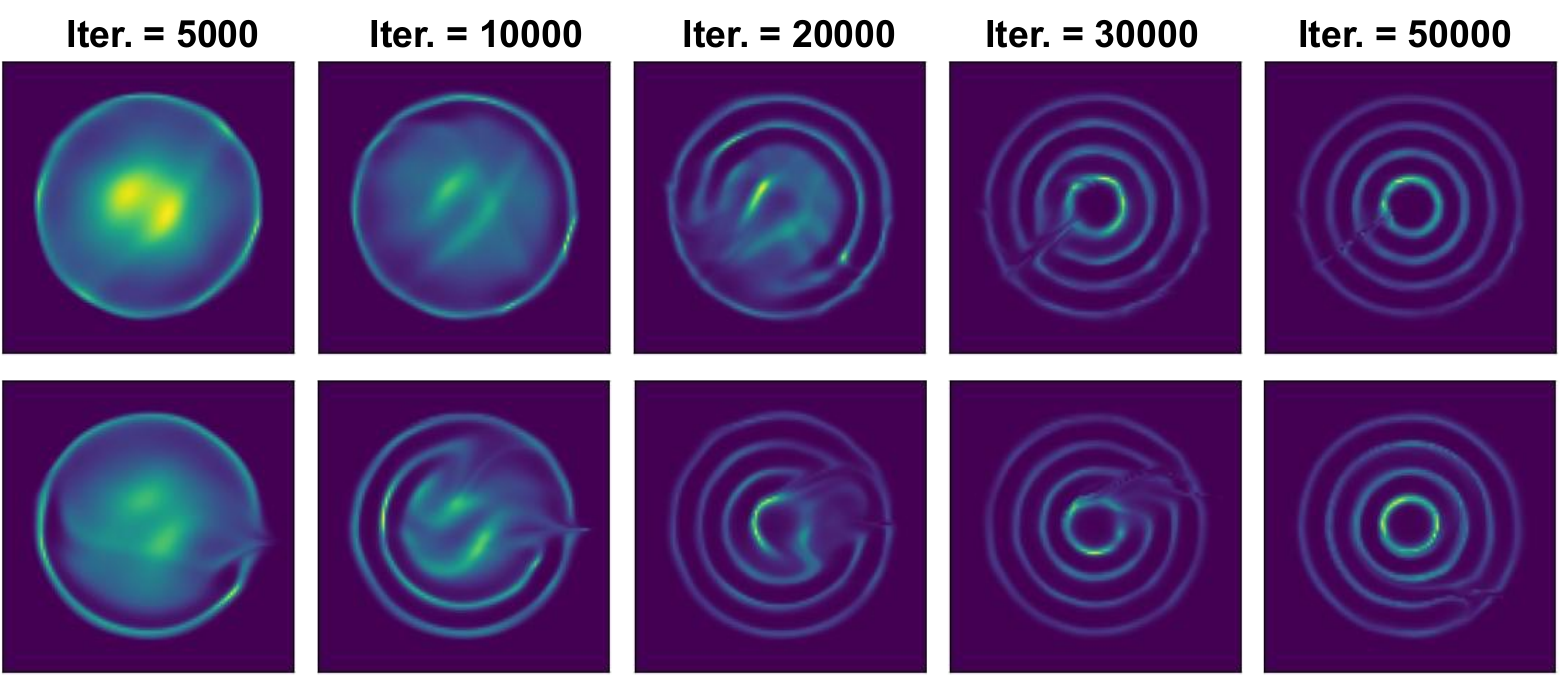}
        \caption{circles}
        \label{fig: cb}
    \end{subfigure}
    
    \caption{\small Visualization of density estimation obtained by \textbf{(Top)} FFJORD and \textbf{(Bottom)} FJORD++ during the training phase on four toy distributions in each sub-figure. The proposed variance reduced model, FJORD++, improves both convergence and training stability, resulting in higher-quality estimated densities.}
    \label{fig: ap_vis_train}
\end{figure}

\begin{figure}[!t]
    \centering
    \begin{subfigure}[b]{0.9\textwidth}
        \centering
        \includegraphics[width=\textwidth]{fig/2s.pdf}
    \end{subfigure}
    \begin{subfigure}[b]{0.9\textwidth}
        \centering
        \includegraphics[width=\textwidth]{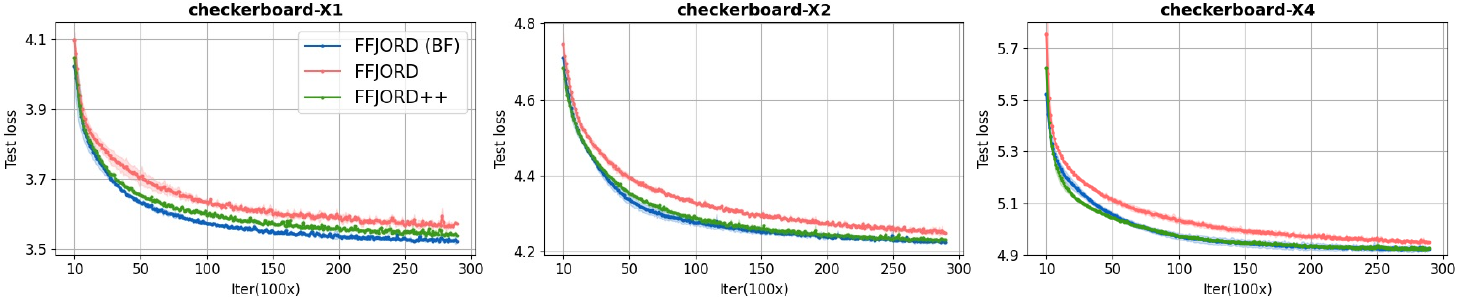}
    \end{subfigure}
    \begin{subfigure}[b]{0.9\textwidth}
        \centering
        \includegraphics[width=\textwidth]{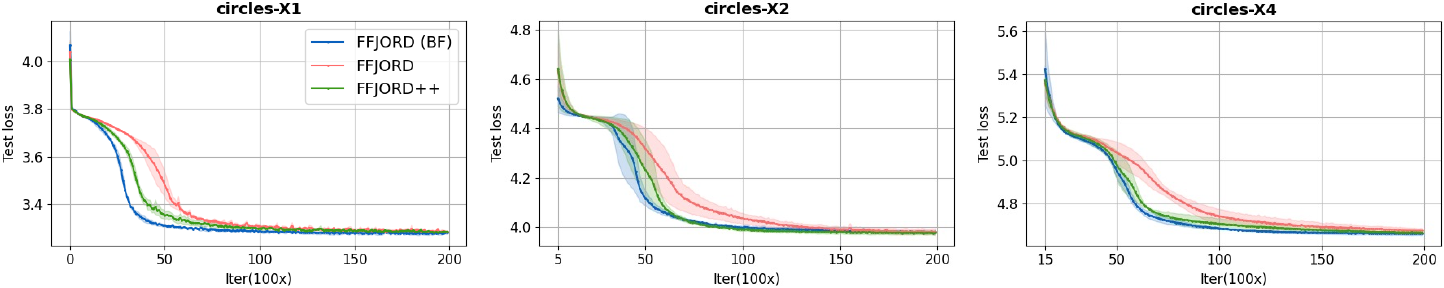}
    \end{subfigure}

    \caption{\small Comparison between the FFJORD++ and FFJORD on data from three distinct shapes with various scales. \textbf{From top to bottom}, each column corresponds to \texttt{2spirals} and \texttt{checkerboard} and \texttt{circles}, respectively. \textbf{From left to right}, each row represents the data with a default scale, as well as shapes stretched 2x and 4x along the X-axis, respectively. As the scale along the X-axis increases, FFJORD++ consistently exhibits superior convergence rates, further widening the performance gap with FFJORD.}

    \label{fig: ap_condition_num}
\end{figure}

\Cref{fig: ap_vis_train} visualizes the comparison of the density estimated by FFJORD and FFJORD++ during the training stage on \texttt{2spirals}, \texttt{checkerboard}, \texttt{rings} and \texttt{circles}.
We provide additional results of comparison between the FFJORD++ and FFJORD on three distinctive shapes (\texttt{2spirals}, \texttt{checkerboard} and \texttt{circles}) with various scales in \Cref{fig: ap_condition_num}.

\subsection{Time Series Modeling}
\begin{figure}[!t]
    \centering
    \includegraphics[width=\linewidth]{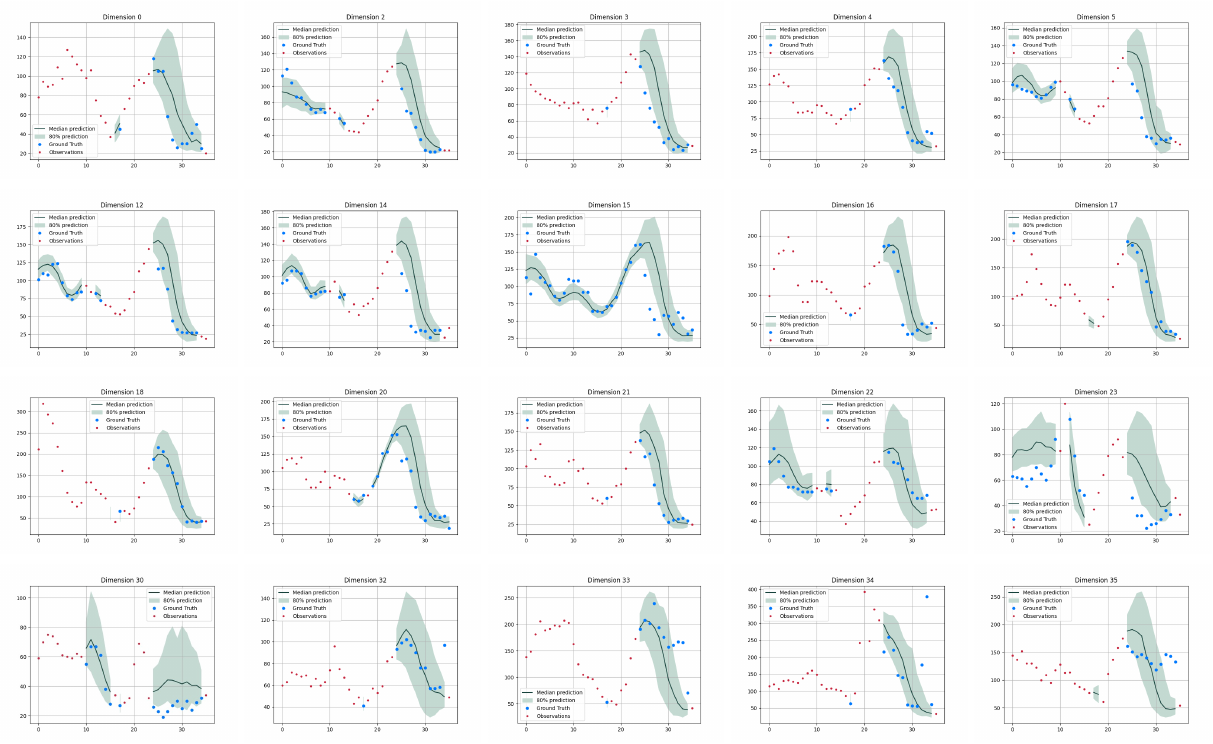}
    \caption{\small \textbf{Imputation} examples for \texttt{PM25} for 20 (out of 36) dimensions.}
    \label{fig: ap_im_pm25}
\end{figure}

\begin{figure}[!t]
    \centering
    \includegraphics[width=\linewidth]{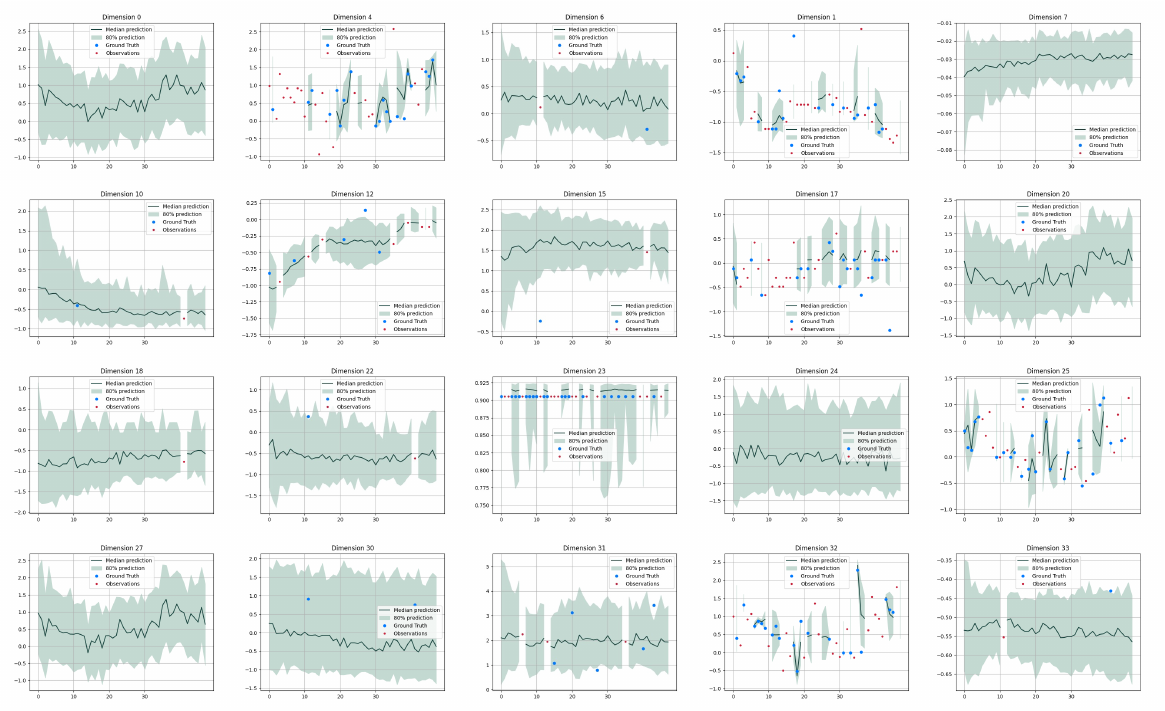}
    \caption{\small \textbf{Imputation} examples for \texttt{Physio 0.5} for 20 (out of 36) dimensions.}
    \label{fig: ap_im_phy}
\end{figure}

\begin{figure}[!t]
    \centering
    \includegraphics[width=\linewidth]{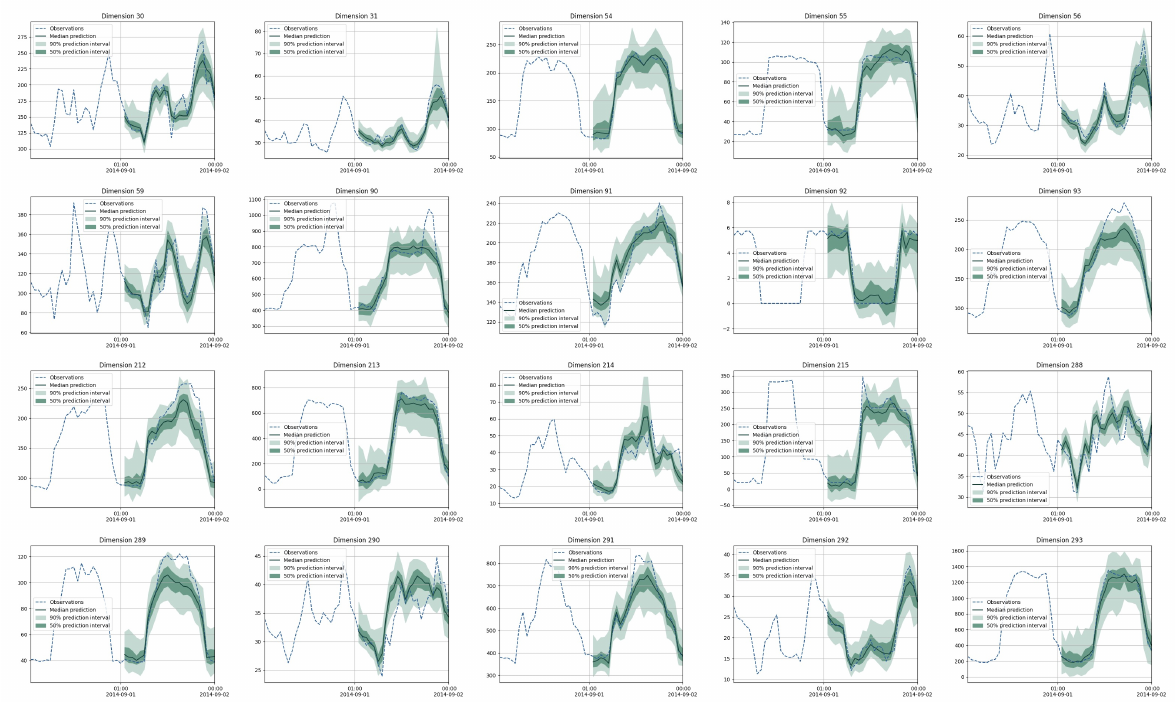}
    \caption{\small \textbf{Forecasting} examples for \texttt{Electricity} for 20 (out of 370) dimensions}
    \label{fig: ap_fore_elec}
\end{figure}

\begin{figure}[!t]
    \centering
    \includegraphics[width=\linewidth]{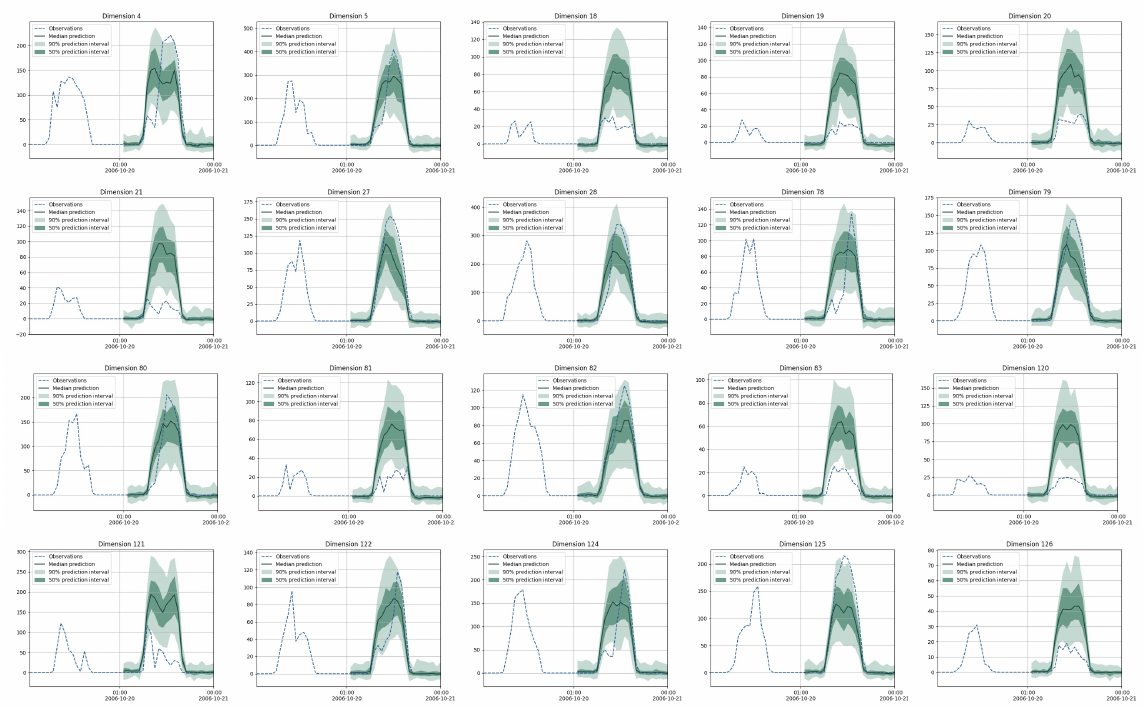}
    \caption{\small \textbf{Forecasting} examples for \texttt{Solar} for 20 (out of 370) dimensions}
    \label{fig: ap_fore_solar}
\end{figure}

\Cref{fig: ap_im_pm25} and \Cref{fig: ap_im_phy} show time series imputation examples obtained by CSBI++ for 20 dimensions on \texttt{PM25} and \texttt{Physio 0.5}, respectively.
\Cref{fig: ap_fore_elec} and \Cref{fig: ap_fore_solar} show time series forecasting examples obtained by CSBI++ for 20 dimensions on \texttt{Electricity} and \texttt{Solar}, respectively.



\subsection{Image Data Modeling}
\begin{figure*}[!t]
    \centering
    \begin{subfigure}[b]{0.9\textwidth}
        \centering
        \includegraphics[width=\textwidth]{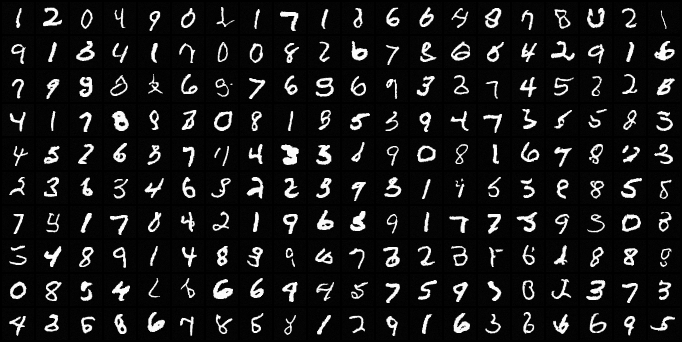}
    \end{subfigure}
    \begin{subfigure}[b]{0.9\textwidth}
        \centering
        \includegraphics[width=\textwidth]{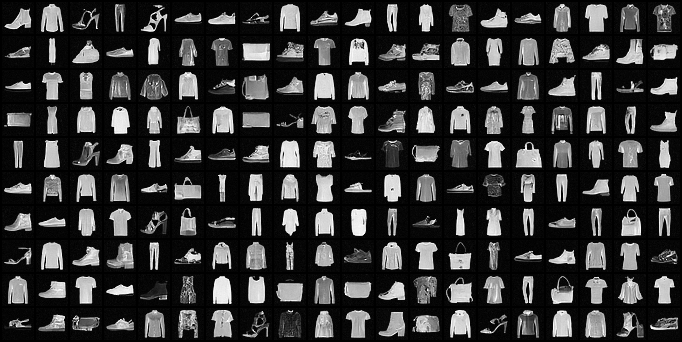}
    \end{subfigure}
    \begin{subfigure}[b]{0.9\textwidth}
        \centering
        \includegraphics[width=\textwidth]{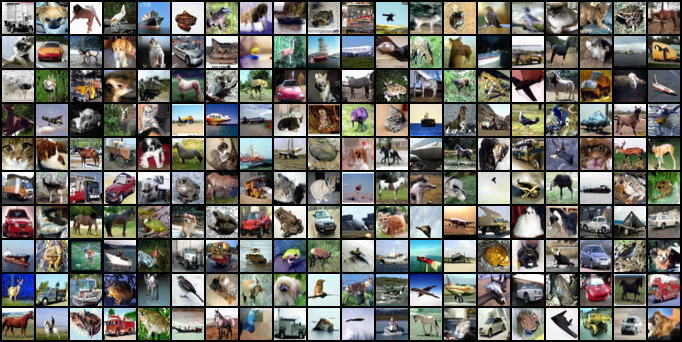}
    \end{subfigure}
    \caption{\small Uncurated samples generated by SB-FBSDE++ on \textbf{(Top)} \texttt{MNIST}, \textbf{(Middle)} \texttt{Fashion-MNIST} and \textbf{(Bottom)} \texttt{CIFAR10}}

    \label{fig: ap_image}
\end{figure*}

We provide images generated by SB-FBSDE++ on three datasets with different patterns in \Cref{fig: ap_image}.

\end{document}